\documentclass{article} % For LaTeX2e
\usepackage{iclr2027_conference,times}

\usepackage{amsmath,amsfonts,bm}

\def\eqref#1{equation~\ref{#1}}
\def\1{\bm{1}}

\DeclareMathAlphabet{\mathsfit}{\encodingdefault}{\sfdefault}{m}{sl}
\SetMathAlphabet{\mathsfit}{bold}{\encodingdefault}{\sfdefault}{bx}{n}

\usepackage{hyperref}
\hypersetup{hidelinks}
\usepackage{url}
\usepackage{graphicx}

\usepackage{booktabs}
\usepackage{multirow}
\usepackage[normalem]{ulem}
\usepackage[table]{xcolor}
\title{PixReenact: Pixel-Conditioned Causal Video Diffusion for Streaming Head-Avatar Reenactment}

\author{%
\makebox[\textwidth][c]{%
\begin{tabular}{c}
\textbf{
Gavriel Habib\textsuperscript{1},
Dvir Samuel\textsuperscript{2},
Or Shimshi\textsuperscript{1},
Rami Ben-Ari\textsuperscript{1}
}
\\[0.35em]
\normalfont
\textsuperscript{1}OriginAI, Israel
\qquad
\textsuperscript{2}NVIDIA
\\[1.2em]
\textbf{Project page:}
\href{https://gavriel-habib.github.io/pixreenact/}
{\texttt{gavriel-habib.github.io/pixreenact/}}
\end{tabular}%
}%
}

\iclrfinalcopy % Uncomment for camera-ready version, but NOT for submission.
\begin{document}

\maketitle

% arXiv preprint: remove ICLR publication header
\fancyhead{}
\renewcommand{\headrulewidth}{0pt}

\begin{abstract}
Streaming head-avatar reenactment aims to animate a reference image according to a live driving video, requiring robust motion transfer, long-term identity stability, and low latency. Existing methods often rely on specialized identity or motion representations, which can discard useful visual information and inherit failure modes from external extractors. In addition, many recent diffusion-based reenactment methods use offline, clip-based generation, jointly processing and denoising an entire video clip before producing its output, making continuous low-latency streaming difficult. We introduce \textbf{PixReenact}, a pixel-conditioned streaming reenactment framework built on causal video diffusion. PixReenact conditions directly on VAE-encoded reference and driving frames, without specialized identity or motion representations. To separate reference identity from driver motion, we train with cross-identity pseudo supervision together with corrective objectives anchored to the original reference and driving inputs. Long self-rollouts reduce autoregressive drift, while state-aware dual-teacher distillation separately addresses cold-start and steady-state generation. Across three cross-identity benchmarks and a long-horizon streaming benchmark, PixReenact demonstrates robust cross-identity reenactment, particularly under challenging conditions such as extreme viewpoints, occlusions, and pronounced facial expressions, while maintaining the reference identity over long streams. A 4-NFE rolling student continuously emits four frames per update with a mean emission latency of 239 ms.
\end{abstract}    
\section{Introduction}
\label{sec:intro}

A streaming head-avatar or portrait-animation continuously animates a reference image from a live driving video, transferring the driver's facial expressions and head motion while preserving the reference identity~\cite{personalive}. Such avatars provide a visual interface for interactive digital agents, with applications in video conferencing, telepresence, and conversational interaction~\cite{app_conversation,app_teleconferencing,talkinghead1kh}. Interactive use requires \textbf{low-latency, stable, and robust streaming reenactment}.

Many existing head-avatar methods condition generation on specialized identity or motion representations, such as implicit keypoints~\cite{liveportrait,personalive}, 3DMM or pose/expression parameters~\cite{GAGAvatar}, and learned identity or motion representations~\cite{x-nemo,Hunyuanportrait,wan-animate}. These representations provide effective and controllable conditioning. Moreover, motion representations that suppress appearance reduce identity information carried by the driving signal, supporting cross-identity reenactment. However, they can discard fine-grained visual information. In particular, methods that rely on external keypoint or motion extractors inherit their failure modes: under severe occlusions, challenging expressions, or large head-pose variations, inaccurate motion estimates can degrade the reenactment. This raises a natural question: can head-avatar reenactment instead operate directly on the visual representations of the reference and driving frames, \textit{without specialized identity or motion representations for conditioning}?

Beyond robust conditioning, a second challenge is low-latency continuous generation. Many recent diffusion-based reenactment methods use offline, clip-based generation, jointly processing and denoising a full video clip before producing output~\cite{wan-animate,Hunyuanportrait,
stable_video_driven_portraits,high_fidelity_relightable}. Consequently, the first output frames become available only after denoising the full clip, making low-latency streaming difficult. Longer videos require processing successive clips, often using overlapping or historical frames to maintain continuity. Such clip-wise continuation may suffer from temporal inconsistency and identity drift. Causal autoregressive video diffusion therefore provides a natural basis for incremental streaming generation.

Motivated by recent advances in causal autoregressive video generation
~\cite{self_forcing,rolling_forcing,samuel2026fast}, we start from a compact
bidirectional video-diffusion backbone and adapt it to causal generation using
the CausVid~\cite{causvid} framework, conditioning directly on VAE-encoded
reference and driving frames. This yields a streaming model without specialized identity or
motion representations. However, this design introduces three challenges. First, under standard self-driving training, direct visual conditioning creates a trivial shortcut: since the reference and driver share the same identity, the model can copy the driver's appearance rather than learn to separate identity from motion. This motivates cross-identity supervision, for which real paired ground truth is generally unavailable. While pseudo targets from an offline reenactment model provide a practical alternative, they may contain identity, motion, or appearance inaccuracies. Second, autoregressive generation repeatedly conditions on its own previous outputs, allowing small errors to accumulate over long streams and cause temporal inconsistency or identity drift. Finally, stream initialization is challenging when the initial driver pose or expression differs substantially from the reference image, requiring a large motion transition while preserving the reference identity. 

To address these challenges, we introduce \textbf{PixReenact}, a pixel-conditioned streaming head-avatar framework built on causal video diffusion. We train PixReenact in a teacher-student framework with cross-identity pseudo pairs generated offline by Wan-Animate~\cite{wan-animate}, together with real self-driving samples for clean reconstruction supervision. Since pseudo targets may contain identity, motion, or appearance inaccuracies, we mitigate their effect using corrective objectives anchored to the reference and driving inputs. To reduce autoregressive drift, the student is trained on long self-rollouts, using its own generated history as at inference. We treat stream initialization separately using state-aware dual-teacher distillation, with cold-start and steady-state teachers. The resulting model continuously emits frames with low latency while conditioning directly on visual inputs.

We evaluate PixReenact across three cross-identity and a long-horizon streaming benchmarks. PixReenact achieves robust cross-identity motion transfer, particularly under challenging conditions such as extreme viewpoints and occlusions, while maintaining the reference identity over long streams. Using a 4-NFE rolling denoising schedule, PixReenact continuously emits four frames per update with a mean emission latency of 239~ms. Our main contributions are:

\begin{itemize}
    \item We introduce a \textbf{pixel-driven streaming reenactment framework} that conditions directly on VAE-encoded reference and driving frames, without specialized identity or motion representations for conditioning.

    \item We develop a \textbf{training strategy for robust cross-identity streaming}, combining cross-identity pseudo supervision, corrective objectives, and long self-rollouts to mitigate driver-appearance leakage and
    long-term autoregressive drift.
    
    \item We propose state-aware dual-teacher distillation, using separate cold-start and steady-state teachers to distill a 4-NFE causal student for low-latency continuous generation.
\end{itemize}

\section{Related Work}
\label{sec:related_work}

\subsection{Head Avatar Reenactment and Conditioning}

Head avatar reenactment animates a reference identity using motion from a driving video. A common strategy is to  transform driving frames into specialized conditioning representations. LivePortrait~\cite{liveportrait} and PersonaLive~\cite{personalive} use implicit keypoint-based motion conditioning, while HunyuanPortrait~\cite{Hunyuanportrait} uses pretrained encoders for implicit motion features. Wan-Animate~\cite{wan-animate} encodes face images into frame-wise implicit latents injected into the diffusion backbone through a dedicated face-attention module. MegActor-$\Sigma$~\cite{megactor_sigma} uses a learned driven encoder to transform the driving video into specialized conditioning features before the denoising transformer. X-NeMo~\cite{x-nemo} learns a 1D identity-agnostic latent motion descriptor from the driving image without pretrained motion detectors, but compresses the driving signal into a specialized latent representation before conditioning the diffusion model.

Other methods explicitly decompose the driving signal: DeX-Portrait~\cite{dexportrait} separates motion into geometric head pose and a learned facial-expression latent. PortraitDirector~\cite{portraitdirector} decomposes the driving signal into specialized pose, local facial-motion, and emotion representations before conditioning the diffusion model. FlashPortrait~\cite{flashportrait} similarly relies on identity-agnostic facial expression features. These representations provide effective motion control and help suppress driver appearance leakage, but their compression may discard fine-grained visual information. Methods relying on external motion extractors can also  inherit their estimation errors.

An alternative retains richer driving-pixel information. MegActor~\cite{megactor} explores raw-video-driven portrait animation and identifies identity leakage as a key challenge of direct visual conditioning. Unlike PixReenact, it uses an image-diffusion backbone with temporal modules, whereas we use a native video-diffusion backbone adapted for causal generation. X-Portrait~\cite{x-portrait} likewise learns motion control from driving RGB and reduces appearance leakage through cross-identity augmentation that modifies the conditioning signal. In contrast, our conditioning remains entirely real, and only the supervision target is synthesized, preserving the same real-driving interface used at inference. Concurrent Wan-Animate-2~\cite{wan-animate-2} similarly uses direct visual conditioning and Wan-Animate-generated cross-identity supervision, whose Lite variant supports autoregressive streaming. Unlike Wan-Animate-2 Lite, PixReenact combines pseudo supervision with reference- and driver-anchored corrective losses during student training, uses separate cold-start and steady-state score teachers, and emits four-frame rather than eight-frame updates.

\subsection{Streaming and Causal Video Generation}

Many recent diffusion-based portrait animation methods target offline generation rather than causal streaming. FlashPortrait~\cite{flashportrait} supports long, potentially unbounded animation through overlapping sliding-window generation and accelerates window-wise denoising by up to $6\times$, but remains offline and relies on extracted facial-expression features. PersonaLive~\cite{personalive} introduces autoregressive micro-chunk generation for low-latency long-term streaming, with sliding training and historical keyframes, while retaining specialized facial-motion and implicit-keypoint conditioning. PortraitDirector~\cite{portraitdirector} supports real-time streaming through diffusion distillation and causal attention, but conditions on decomposed pose, local facial-motion, and emotion representations. In contrast, PixReenact combines low-latency causal streaming with direct visual conditioning rather than specialized identity or motion representations.

Beyond portrait animation, recent causal video-diffusion works~\cite{causvid,self_forcing,rolling_forcing,causal_forcing} show that video diffusion models can be adapted and distilled into few-step autoregressive generators for low-latency streaming. These works establish causal autoregressive video diffusion as a practical approach for streaming, but do not address pixel-conditioned cross-identity reenactment challenges, including driver-appearance leakage, imperfect cross-identity supervision, and large reference-to-driver transitions at stream initialization. PixReenact builds on this paradigm with training mechanisms tailored to these challenges.
\section{Method}
\label{sec:method}

\subsection{Overview}

Head avatar reenactment animates a reference image \(I_{\mathrm{ref}}\) using the facial expression, head pose, and motion of a driving video \(D_{1:T}\). We consider the challenging setting of \emph{streaming head-avatar reenactment}, where driving frames arrive sequentially and the model must continuously generate \(Y_{1:T}\), preserving the reference identity while transferring the driver's motion with low latency. Unlike methods relying on specialized conditioning representations such as keypoints or learned motion codes that may discard fine-grained visual cues, PixReenact conditions directly on VAE-encoded reference and driving frames. Avoiding such bottlenecks is important in streaming, since lost information cannot be recovered and errors may propagate through the autoregressive context.

Direct visual conditioning introduces challenges for cross-identity reenactment and autoregressive streaming. We therefore combine cross-identity supervision with corrective objectives to reduce appearance leakage, train on long self-rollouts to reduce autoregressive drift, and treat stream initialization separately from steady-state generation. At inference, a causal rolling denoising process continuously incorporates new driving frames and emits new video frames with low latency.

\begin{figure}[t]
    \centering
    \includegraphics[width=\textwidth]{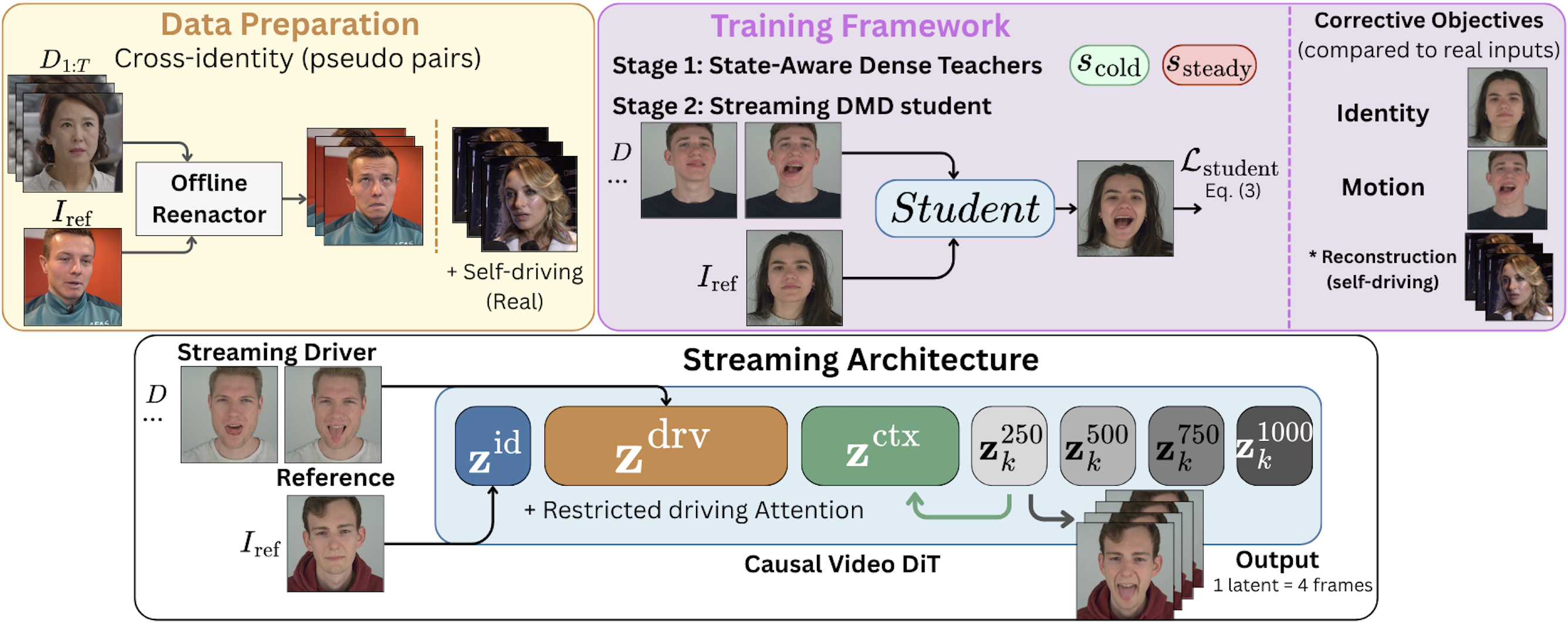} \caption{\textbf{PixReenact Overview.} Top-left: offline cross-identity pseudo targets from real reference-driver pairs, plus real self-driving data. Top-right: cold-start and steady-state teachers are distilled into a 4-NFE student with DMD and corrective objectives. Bottom: reference, driving, and context latents condition a causal video DiT with restricted driving attention and rolling four-frame emission.}
    \label{fig:method_overview}
\end{figure}

\subsection{From Video Diffusion to Pixel-Driven Streaming Reenactment}
\label{subsec:pixel_driven}

A natural route to streaming reenactment is adapting a diffusion-based reenactment model for causal generation. However, such models typically target high-quality offline generation and are not
naturally suited to low-latency streaming~\cite{personalive,flashportrait}. We instead convert a compact bidirectional video-diffusion backbone into a causal autoregressive
generator using the framework of~\cite{causvid}. This yields an efficient streaming backbone, but not yet a reenactment model.

To obtain robust reenactment without specialized conditioning, we
condition directly on VAE-encoded reference and driving frames. Figure~\ref{fig:method_overview} (bottom) shows the resulting
streaming architecture. Let $\mathcal{E}$ and $\mathcal{D}$ denote the frozen VAE encoder and decoder. Given a reference image $I_{\mathrm{ref}}$ and driving frames $D_{1:T}$, we obtain an identity latent $\mathbf{z}^{\mathrm{id}}=\mathcal{E}(I_{\mathrm{ref}})$ and driving latents $\mathbf{z}^{\mathrm{drv}}_{1:T'}=\mathcal{E}(D_{1:T})$. The causal VAE compresses time by four, so each latent spans four frames and $T'=\lceil T/4\rceil$. All temporal indices below refer to this latent grid. The reference and driving latents become DiT tokens and are concatenated with the context and video tokens into a shared self-attention sequence.

At streaming iteration $k$, the model processes
\[
[\mathbf{z}^{\mathrm{id}} \;|\; \mathbf{z}^{\mathrm{drv}} \;|\; \mathbf{z}^{\mathrm{ctx}} \;|\; \mathbf{z}_k^{250},\mathbf{z}_k^{500},\mathbf{z}_k^{750},\mathbf{z}_k^{1000}],
\]
where $\mathbf{z}^{\mathrm{drv}}$ denotes the local driving latents used for the current generation window, and $\mathbf{z}^{\mathrm{ctx}}$ contains two slots filled by the latest
emitted video latents. Before sufficient history is available, context
slots are initialized from the reference. The term $\mathbf{z}_k^{t}$ denotes
an in-flight video latent at diffusion timestep $t$; all in-flight latents are indexed by the current iteration for brevity. At each iteration, a new latent enters at $t=1000$. The forward
pass advances all in-flight latents by one level and emits the cleanest as
$\hat{\mathbf{z}}_k$. A latent admitted at iteration $k$ undergoes four
passes and is emitted at $k{+}3$. It is decoded into four output frames,
$Y_k=\mathcal{D}(\hat{\mathbf{z}}_k)$, using a causal streaming VAE decoder
with a persistent temporal cache. The latest emitted latents update
$\mathbf{z}^{\mathrm{ctx}}$ for subsequent iterations. We obtain the causal
model by applying causal attention masks to the bidirectional backbone and
adapt it using self-attention LoRA. The rolling denoising and KV-cache design
follows~\cite{causvid}.

To preserve temporal correspondence, the video latent at position $n$ attends
only to a three-latent driving window $\mathbf{z}^{\mathrm{drv}}_{n-1:n+1}$ through a restricted attention mask. Neighboring latents
provide temporal context around the aligned driving latent while
preventing attention to unrelated positions. Although this includes one-latent lookahead, it adds no latency beyond the rolling denoising pipeline: the four-step staircase ensures
$\mathbf{z}^{\mathrm{drv}}_{n+1}$ is available before latent $n$ is emitted.
Thus, the model gains local temporal context while retaining streaming
generation. We evaluate this window in
Sec.~\ref{subsec:ablations}.

However, because the driving latents also contain the driver's appearance,
standard self-driving training creates a strong shortcut. Since
$I_{\mathrm{ref}}$ and $D_{1:T}$ share the same identity, the model is not
explicitly required to preserve identity from $I_{\mathrm{ref}}$ while
transferring pose, expression, and motion from $D_{1:T}$. It can instead copy
appearance directly from the driving stream.

\subsection{Learning Cross-Identity Reenactment}
To remove this shortcut, we use \textbf{cross-identity supervision},
where reference and driving inputs come from different identities.
Cross-identity training has also been explored in prior portrait-animation
methods using synthetic, retargeted, or augmented pairs~\cite{x-portrait,dreamactorm2,megactor,scail2}. Since real paired
cross-identity ground truth with matched motion is generally unavailable, we
construct pseudo pairs using a strong offline
reenactment model~\cite{wan-animate}, as illustrated in Figure~\ref{fig:method_overview} (top-left). We sample a reference image $I_{\mathrm{ref}}^{A}$ and driving video
$D_{1:T}^{B}$ from different identities ($A \neq B$), and use the offline model to generate a pseudo target $\tilde{Y}_{1:T}^{A \leftarrow B}$ intended to preserve the appearance of $A$ while following the motion of $B$. Unlike prior strategies that modify or synthesize the cross-identity
conditioning signal, our reference and driving inputs remain entirely real,
while only the supervision target is synthesized. Thus, training uses the same real-driving input interface as inference.

Pseudo targets may contain identity, pose, expression, and visual-quality
errors. We therefore filter low-quality targets using identity, motion, and
emotion criteria (Appendix~\ref{subsec:supp_data}). During student training,
corrective objectives use the original inputs: identity is supervised against
$I_{\mathrm{ref}}^{A}$, while pose, expression, and emotion are supervised
against the real driver $D_{1:T}^{B}$
(Sec.~\ref{subsec:corrective_training_objectives}). We also mix real
self-driving samples for clean reconstruction and reduced reliance on imperfect
pseudo targets. Together, these signals retain the benefits of cross-identity
supervision while mitigating pseudo-target errors.

\subsection{State-Aware Streaming Distillation}
\label{subsec:streaming_distillation}

While the above training enables cross-identity reenactment, we next describe how to achieve low-latency streaming. As shown in
Figure~\ref{fig:method_overview} (top-right), we first train causal dense-timestep score teachers from the bidirectional backbone, then distill
them into a causal 4-NFE streaming student using Distribution Matching
Distillation (DMD)~\cite{DMD}, following~\cite{causvid}. Teachers and student share the same causal architecture and pixel-conditioning interface.

\textbf{Stage 1: State-aware dense teachers.} We apply the causal attention masks of Sec.~\ref{subsec:pixel_driven} to the bidirectional backbone and train self-attention LoRA over a dense range of noise levels using cross-identity supervision.
The first streaming emission differs substantially from later ones. At $k=1$, no generated history is available, so
$\mathbf{z}^{\mathrm{ctx}}$ contains only reference-initialized latents. This
creates a challenging \textbf{cold-start} state, especially when reference and driver have a large pose gap: the model must establish the
driver's pose while preserving the reference identity. For $k>1$, the context contains generated history already better aligned with the current driver state, so the model mainly follows
incremental motion while maintaining temporal consistency and identity.

We therefore train two dense teachers specialized to these rollout states.
The cold-start teacher uses reference-initialized context and emphasizes large
reference-driver pose gaps, whereas the steady-state teacher uses preceding
target latents as context, matching later rollouts. Both teachers retain the
base video-diffusion model's native flow-velocity parameterization, but are
supervised using MSE on the implied clean-latent estimate. For each clean
target latent $\mathbf{z}^{0}$, we independently sample a diffusion timestep $t$,
while keeping context latents clean ($t=0$), construct $\mathbf{z}^{t}$, and
obtain $\hat{\mathbf{z}}^{0}=\mathbf{z}^{t}-\sigma_t\hat{\mathbf{v}}^{t}$.
The predicted flow is also converted to the corresponding score for
distillation. The real-score teacher used during distillation is

\begin{equation}
s_{\mathrm{real}}^{(k)} = s_{\mathrm{cold}}\ \text{if } k=1,\quad
s_{\mathrm{steady}}\ \text{if } k>1.
\end{equation}

\textbf{Stage 2: Streaming DMD student.}
The student follows the 4-level denoising staircase as at inference
(Sec.~\ref{subsec:pixel_driven}), and is trained with long self-rollouts. This exposes it to its own history,
reducing training-inference mismatch and long-horizon error accumulation.

We re-noise an emitted latent $\hat{\mathbf{z}}_k$ at timestep $t$ to obtain
$\mathbf{z}_k^t$. The frozen state-dependent teacher
$s_{\mathrm{real}}^{(k)}$ estimates the target score, while the online
$s_{\mathrm{fake}}$ estimates the student score. The DMD gradient is
\begin{equation}
\label{eq:dmd}
\mathbf{g}_{\mathrm{DMD}}^{(k)}
= w(t)\!\left[s_{\mathrm{fake}}(\mathbf{z}_k^t,t)
- s_{\mathrm{real}}^{(k)}(\mathbf{z}_k^t,t)\right],
\end{equation}
where $w(t)$ is the standard DMD timestep weight and both scores condition on
the identity, driving, and context tokens. We use the standard surrogate
$\mathcal{L}_{\mathrm{DMD}}=\tfrac{1}{2}\|\hat{\mathbf{z}}_k-
\operatorname{sg}(\hat{\mathbf{z}}_k-\mathbf{g}_{\mathrm{DMD}}^{(k)})\|_2^2$,
with gradient $\mathbf{g}_{\mathrm{DMD}}^{(k)}$ w.r.t.
$\hat{\mathbf{z}}_k$, backpropagated through the rollout. See
Appendix~\ref{subsec:supp_teachers},~\ref{subsec:supp_distill} for training details.

\subsection{Corrective Training Objectives}
\label{subsec:corrective_training_objectives}

In addition to DMD, we apply corrective objectives during student training.
Cross-identity pseudo targets may contain errors in identity, motion, and
appearance, so these objectives are computed from the original real inputs.
At emission $k$, let $Y_k$ denote the four emitted frames and $D_k$ their
aligned driving frames. Frame-wise losses are averaged over the four frames.

For \textbf{identity preservation}, we use
$\mathcal{L}_{\mathrm{id}}=\max\!\bigl(0,m-\cos(\phi(Y_k),\phi(I_{\mathrm{ref}}))\bigr)$,
where $\phi(\cdot)$ is a frozen identity embedding and $m$ reflects typical
same-identity similarity, avoiding unrealistic pressure. To
suppress driver-identity leakage, we use
$\mathcal{L}_{\mathrm{leak}}=\max\!\bigl(0,\cos(\phi(Y_k),\phi(D_k))-m_{\mathrm{leak}}\bigr)$
on cross-identity samples only. For \textbf{motion supervision}, we apply
$\ell_1$ losses between estimated head-pose and expression parameters of $Y_k$
and driver $D_k$
($\mathcal{L}_{\mathrm{pose}},\mathcal{L}_{\mathrm{expr}}$), rather than the
pseudo target. We also enforce emotion consistency using cosine similarity
between extracted emotion features ($\mathcal{L}_{\mathrm{emo}}$).

For real self-driving samples, exact reconstruction targets are available, so
we apply LPIPS and adversarial losses for visual fidelity. All supervised
samples, real and pseudo, also receive an $\ell_2$ latent anchor
$\mathcal{L}_{\mathrm{anc}}$ toward their target latents. The full student
objective is
\begin{equation}
\begin{aligned}
\mathcal{L}_{\mathrm{student}}
={}&\lambda_{\mathrm{DMD}}\mathcal{L}_{\mathrm{DMD}}
+\lambda_{\mathrm{anc}}\mathcal{L}_{\mathrm{anc}}
+\lambda_{\mathrm{id}}\mathcal{L}_{\mathrm{id}}
+\lambda_{\mathrm{leak}}\mathcal{L}_{\mathrm{leak}} \\
&+\lambda_{\mathrm{pose}}\mathcal{L}_{\mathrm{pose}}
+\lambda_{\mathrm{expr}}\mathcal{L}_{\mathrm{expr}}
+\lambda_{\mathrm{emo}}\mathcal{L}_{\mathrm{emo}}
+\lambda_{\mathrm{rec}}\mathcal{L}_{\mathrm{rec}},
\end{aligned}
\end{equation}
where $\mathcal{L}_{\mathrm{rec}}$ combines the LPIPS and adversarial losses on
real self-driving samples. Estimator details, margins, loss weights, and
implementation details are provided in Appendix~\ref{subsec:supp_losses}.
\section{Evaluation}
\label{sec:evaluation}
\newcommand{\std}[1]{\,{\scriptsize$\pm$#1}}

\subsection{Experimental Setup}
\label{subsec:experimental_setup}

\noindent\textbf{Training data.}
Following PersonaLive~\cite{personalive}, we use the same three datasets for self-driving training: NeRSemble~\cite{nersemble}, VFHQ~\cite{vfhq}, and DH-FaceVid-1K~\cite{dhfacevid}. PersonaLive does not report its number of training clips. Our self-driving set contains 1{,}548 clips: 402 from NeRSemble, 146 from VFHQ, and 1{,}000 from DH-FaceVid-1K. For cross-identity training, we reuse these clips with different-identity references and add 800 FFHQ~\cite{ffhq} references, each paired with three drivers. Wan-Animate~\cite{wan-animate} generates offline pseudo targets, yielding 13{,}561 pairs after label-quality filtering (Sec.~\ref{subsec:supp_data}).

All clips are 4-second, $512^2$ segments. For self-driving, the first frame is the reference and the clip both driver and ground truth. We call the Wan-Animate-supervised model \textbf{PixReenact}. To test pseudo-label
sensitivity, we train an otherwise identical \emph{PL-labels} variant using
PersonaLive~\cite{personalive} with the same reference-driver pairings.
Details and results are in the Appendix
(Sec.~\ref{subsec:supp_data},~\ref{sec:supp_additional_results}).

Training and evaluation sets are strictly disjoint. NeRSemble evaluation uses 15 held-out subjects, VFHQ uses VFHQ-Test, and the 50 FFHQ evaluation images are not used in training. TalkingHead-1KH~\cite{talkinghead1kh}
and CelebV-HQ~\cite{celebvhq} are used only for evaluation.

\noindent\textbf{Baselines and metrics.}
We compare PixReenact with image-based methods
LivePortrait~\cite{liveportrait} and X-Portrait~\cite{x-portrait}; offline
video methods MegActor-$\Sigma$~\cite{megactor_sigma},
Wan-Animate~\cite{wan-animate}, X-NeMo~\cite{x-nemo},
Wan-Animate-2-Base~\cite{wan-animate-2} (Wan-Animate-2 in all tables), and
FlashPortrait~\cite{flashportrait}; and  streaming method PersonaLive~\cite{personalive}. All baselines are run using their official releases; implementation details are in Appendix~\ref{subsec:supp_baselines}. 
We measure identity preservation by
cosine similarity between AdaFace embeddings~\cite{adaface} of generated
frames and the reference image (ID). Motion transfer uses average expression (AED) and pose (APD) distances from SMIRK~\cite{smirk} estimates for generated and driving frames. Temporal consistency and video quality are measured by tLP~\cite{tLP} and
Fréchet Video Distance (FVD)~\cite{FVD}, respectively. For streaming methods, we report mean steady-state emission latency per
four-frame chunk and throughput in frames per second (FPS),
computed as $4/\text{latency}$. The metrics capture complementary reenactment quality aspects and should be interpreted jointly; their limitations are discussed in Appendix~\ref{sec:metric_discussion}.

\noindent\textbf{Evaluation benchmarks.}
We evaluate PixReenact on three cross-identity protocols and one long-horizon
streaming benchmark. First, we introduce a 68-pair extreme-condition benchmark with CelebV-HQ drivers~\cite{celebvhq} and FFHQ references, targeting
large viewpoint changes and challenging occlusions. We also evaluate on two
standard benchmarks. Following LivePortrait~\cite{liveportrait}, we use 25
TalkingHead-1KH test clips~\cite{talkinghead1kh} still available on YouTube
(out of 35 used in~\cite{liveportrait}) and 50 VFHQ~\cite{vfhq} test clips as
drivers, paired with the first 50 FFHQ images~\cite{ffhq} as references. We
also construct a 70-pair benchmark from 15 held-out NeRSemble
subjects~\cite{nersemble}, with five providing references and all 15 serving
as drivers. For long-horizon streaming, we evaluate identity stability on 28
cross-identity videos over five minutes (details in
Appendix~\ref{subsec:supp_benchmarks}). We additionally report self-reenactment
and single-window cross-reenactment results in the Appendix
(Tables~\ref{tab:public_benchmark_self}, \ref{tab:nersemble_self},
and~\ref{tab:single_window_evaluation}).

\noindent\textbf{Implementation details.}
PixReenact uses Wan2.1-T2V-1.3B~\cite{wan2.1} as its bidirectional video-diffusion backbone
and adopts the causal generation and DMD distillation framework of
CausVid~\cite{causvid}. We apply causal attention masks and use a 4-NFE rolling
denoising schedule with timesteps $[1000,750,500,250,0]$, adapting the DiT with
self-attention LoRA. All experiments use $512^2$ resolution, with runtime
measurements on a single NVIDIA H200 GPU. Additional training
hyperparameters and implementation details are provided in the Appendix~\ref{sec:supp_details}.

\subsection{Robust Streaming Reenactment}
\label{subsec:robust_streaming}
\begin{figure*}[t]
  \centering
  \includegraphics[width=\textwidth]{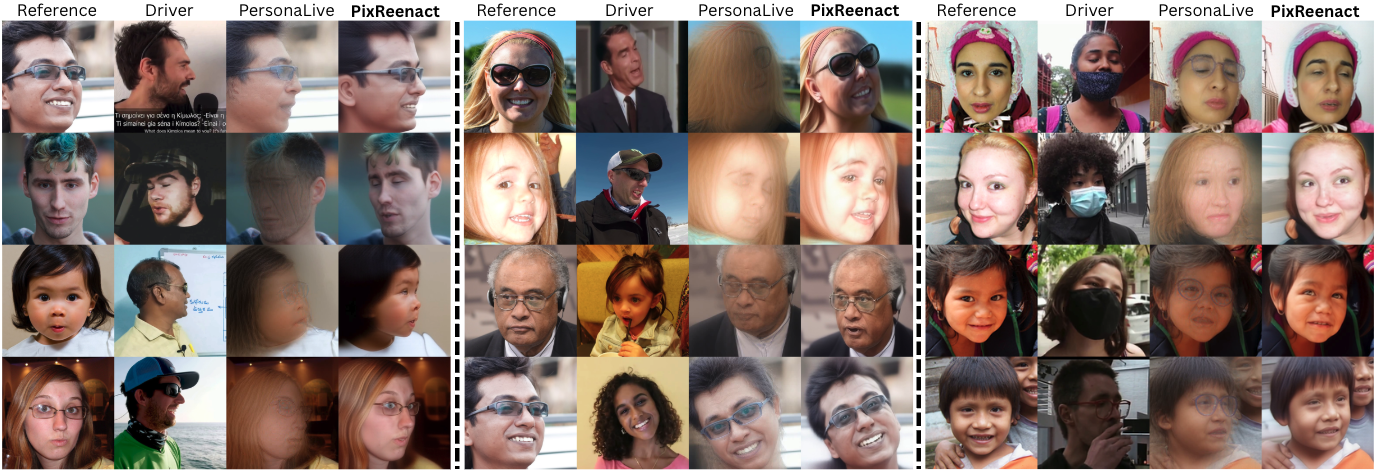}
  \caption{\textbf{Qualitative comparison under challenging viewpoints and occlusions.} Compared with PersonaLive, PixReenact better preserves the reference identity while following the driver's motion.}
\label{fig:qual_extreme}
\end{figure*}

\paragraph{Extreme viewpoints and occlusions.}
We first evaluate the challenging conditions motivating our pixel-driven
formulation, comparing streaming methods on difficult inputs.
Figure~\ref{fig:qual_extreme} shows representative examples with large
viewpoints and occlusions, including hands, face masks, and facial
accessories. PixReenact remains effective while preserving the reference
appearance.

Table~\ref{tab:extreme_benchmark} reports results on our 68-pair
CelebV-HQ extreme-condition benchmark. PixReenact substantially improves
identity preservation over PersonaLive (ID $0.727$ vs.\ $0.472$), lowers AED ($1.135$ vs.\ $1.295$), with similar tLP
($1.853$ vs.\ $1.866$). PersonaLive obtains lower APD and FVD, despite visible
degradation in several qualitative examples. PixReenact's clearest
advantage in this challenging setting is its substantially stronger reference-identity preservation. This highlights the importance of
interpreting metrics jointly under extreme viewpoints and occlusions (see
Appendix~\ref{sec:metric_discussion}).

\begin{table*}[t]
\centering
\caption{\textbf{Extreme-condition benchmark} (68 pairs).
FFHQ references driven by CelebV-HQ clips with extreme poses and
occlusions. Mean$\pm$std over clips; FVD is set-level.}
\label{tab:extreme_benchmark}
\small
\setlength{\tabcolsep}{4pt}
\begin{tabular}{lccccc}
\toprule
Method & ID$\uparrow$ & AED$\downarrow$ & APD$\downarrow$ & tLP$\downarrow$ & FVD$\downarrow$ \\
\midrule
PersonaLive & 0.472\std{0.23} & 1.295\std{0.64} & \textbf{0.135}\std{0.10} & 1.866\std{1.57} & \textbf{643.5} \\
PixReenact & \textbf{0.727}\std{0.11} & \textbf{1.135}\std{0.30} & 0.169\std{0.12} & \textbf{1.853}\std{1.35} & 662.5 \\
\bottomrule
\end{tabular}
\end{table*}

\begin{figure*}[t]
    \centering
    \includegraphics[width=0.3\linewidth]{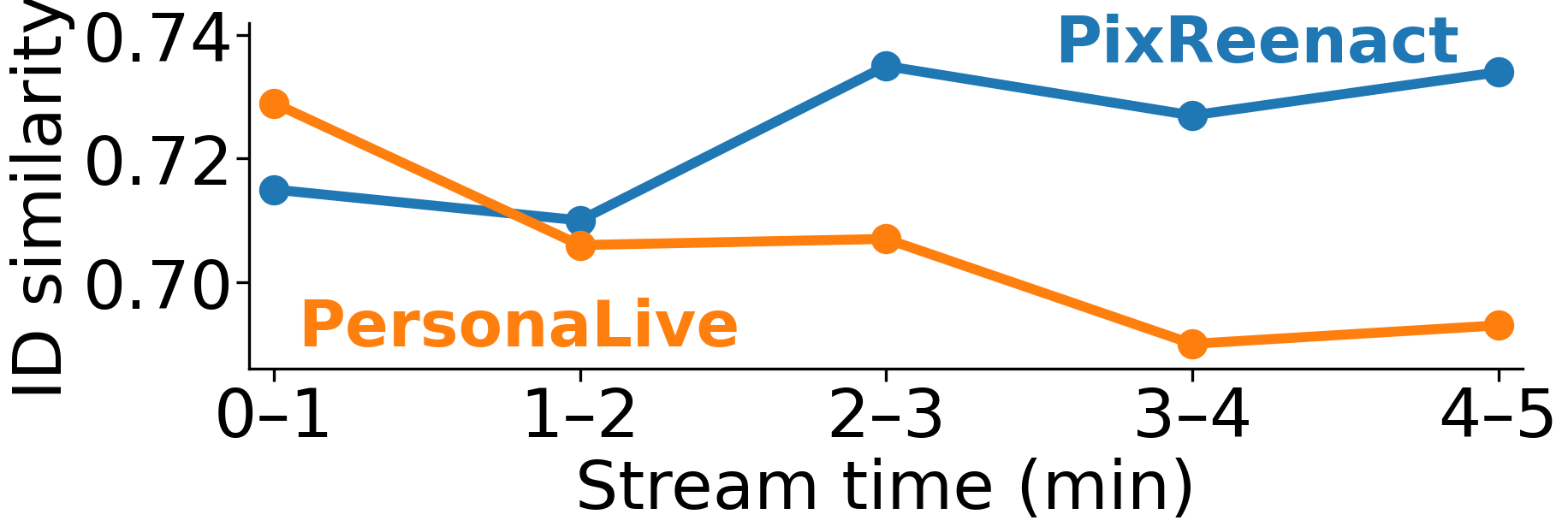}
    \caption{\textbf{Long-term identity stability.}
    AdaFace similarity over five minutes on 28 clips.}
    \label{fig:id_sim_long_videos}
\end{figure*}

\paragraph{Long-horizon streaming.}
Autoregressive self-conditioning can accumulate errors, so we evaluate identity
stability over 28 five-minute clips. Figure~\ref{fig:id_sim_long_videos} reports
mean AdaFace similarity per minute: PersonaLive declines over time, while
PixReenact remains stable. Appendix Figure~\ref{fig:long_horizon_qual} provides
qualitative examples, showing reference-identity stability over horizons much longer than training.

\subsection{Standard Cross-Identity Benchmarks}
\label{subsec:standard_cross_reenactment}
\begin{figure*}[t]
  \centering
  \includegraphics[width=\textwidth]{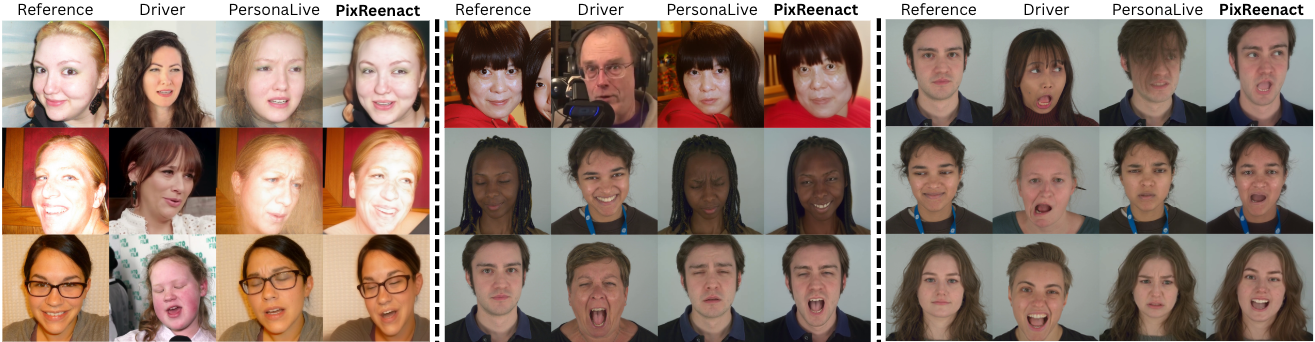}
  \caption{\textbf{Qualitative cross-identity reenactment.} PixReenact better preserves reference identity while transferring driver facial motion on the public and NeRSemble benchmarks.}  
  \label{fig:qual_standard}
\end{figure*}

We next evaluate PixReenact on the public and NeRSemble cross-identity
benchmarks. Figure~\ref{fig:qual_standard} shows qualitative results.
PixReenact preserves reference identity while following the driver's
facial motion, including pronounced expressions. Additional
comparisons against all baselines are shown in Appendix
Figure~\ref{fig:qual_cross_additional}. Table~\ref{tab:cross_benchmarks}
compares image-based, offline video-based, and streaming methods. On the public
benchmark, PixReenact achieves the best FVD (377.1) and better identity
preservation than PersonaLive (ID $0.709$ vs.\ $0.662$), with comparable AED
($0.790$ vs.\ $0.788$). PersonaLive performs better in APD and tLP. On
NeRSemble, PixReenact improves over PersonaLive in ID ($0.753$ vs.\ $0.734$),
AED ($0.714$ vs.\ $0.824$), tLP ($0.803$ vs.\ $0.890$), and FVD ($264.8$ vs.
$310.0$), while remaining close in APD ($0.0426$ vs.\ $0.0392$). Overall,
PixReenact remains competitive with offline methods while streaming
continuously without specialized extractors at inference.

PixReenact emits four frames at 238.9~ms (16.74 FPS), versus 211.1~ms
(18.95 FPS) for PersonaLive. The gap mainly reflects our video-native causal
DiT and causal VAE decoder, compared with PersonaLive's image-based U-Net and
VAE decoder. A detailed comparison is in Appendix Table~\ref{tab:latency_analysis}.

\begin{table*}[t]
\centering
\caption{\textbf{Cross-identity reenactment} on public (75 pairs) and
NeRSemble (70 pairs) benchmarks. Single-H200 runtime; latency is mean steady-state emission time. Better streaming results are bold.}
\label{tab:cross_benchmarks}
\setlength{\tabcolsep}{2.4pt}
\resizebox{\textwidth}{!}{%
\begin{tabular}{llccccc|ccccc|cc}
\toprule
& &
\multicolumn{5}{c|}{\textbf{Public benchmark (75 pairs)}}
& \multicolumn{5}{c|}{\textbf{NeRSemble (70 pairs)}}
& \multicolumn{2}{c}{\textbf{Runtime}} \\
\cmidrule(lr){3-7}
\cmidrule(lr){8-12}
\cmidrule(lr){13-14}
& Method
& ID$\uparrow$ & AED$\downarrow$ & APD$\downarrow$ & tLP$\downarrow$ & FVD$\downarrow$
& ID$\uparrow$ & AED$\downarrow$ & APD$\downarrow$ & tLP$\downarrow$ & FVD$\downarrow$
& Lat.\,$\downarrow$ & FPS$\uparrow$ \\
\midrule

\multirow{2}{*}{\textit{Per-frame}}
& LivePortrait
& 0.863 & 1.071 & 0.208 & 1.146 & 579.0
& 0.849 & 0.787 & 0.0840 & 0.763 & 302.4
& -- & 18.90 \\
& X-Portrait
& 0.701 & 0.887 & 0.097 & 1.613 & 595.2
& 0.774 & 0.694 & 0.0461 & 0.795 & 281.2
& -- & 0.65 \\

\midrule
\multirow{5}{*}{\textit{Offline}}
& MegActor-$\Sigma$
& 0.554 & 0.841 & 0.106 & 1.910 & 582.7
& 0.729 & 0.802 & 0.0566 & 0.976 & 247.4
& -- & 0.49 \\
& Wan-Animate
& 0.570 & 0.966 & 0.089 & 1.903 & 579.2
& 0.632 & 0.789 & 0.0381 & 1.199 & 201.3
& -- & 0.41 \\
& X-NeMo
& 0.614 & 0.775 & 0.065 & 1.378 & 449.1
& 0.710 & 0.639 & 0.0291 & 0.733 & 276.5
& -- & 1.37 \\
& Wan-Animate-2
& 0.603 & 0.863 & 0.085 & 4.500 & 655.8
& 0.583 & 0.723 & 0.0451 & 1.831 & 276.5
& -- & 0.29 \\
& FlashPortrait
& 0.760 & 0.864 & 0.112 & 3.072 & 633.7
& 0.800 & 0.758 & 0.0598 & 1.527 & 303.2
& -- & 3.03 \\

\midrule
\multirow{2}{*}{\textit{Streaming}}
& PersonaLive
& 0.662 & \textbf{0.788} & \textbf{0.050} & \textbf{1.059} & 407.1
& 0.734 & 0.824 & \textbf{0.0392} & 0.890 & 310.0
& \textbf{211.1} & \textbf{18.95} \\
& PixReenact
& \textbf{0.709} & 0.790 & 0.066 & 1.234 & \textbf{377.1}
& \textbf{0.753} & \textbf{0.714} & 0.0426 & \textbf{0.803} & \textbf{264.8}
& 238.9 & 16.74 \\

\bottomrule
\end{tabular}%
}
\end{table*}

\subsection{Ablation Studies}
\label{subsec:ablations}

% % Main-paper ablations for Sec. 4.5 — ordered by method story
% \begin{table*}[t]
% \centering
% \caption{\textbf{Component ablations} on the 75-pair public cross-identity benchmark. All variants continue from the same checkpoint for 5K steps; the first row is the matched full-recipe continuation. IDD measures driver-identity similarity (lower is better). $^{\dagger}$Driver copying can make motion scores appear better without a true improvement, while reference identity degrades (lower ID, higher IDD).}
% \label{tab:ablation_components}
% \small
% \setlength{\tabcolsep}{4pt}
% \begin{tabular}{llcccccc}
% \toprule
% & Variant
% & ID$\uparrow$ & IDD$\downarrow$ & AED$\downarrow$ & APD$\downarrow$
% & tLP$\downarrow$ & FVD$\downarrow$ \\
% \midrule

% & Full recipe (continuation baseline)
% & 0.707 & 0.023 & 0.816 & 0.071 & 1.254 & 389 \\

% \midrule
% \multirow{1}{*}{\textit{Pixel cond.}}
% & Driver attention $w=0$$^{\dagger}$
% & 0.585 & 0.210 & 0.679 & 0.067 & 1.690 & 273 \\

% \midrule
% \multirow{3}{*}{\textit{Cross-ID}}
% & w/o cross-identity supervision
% & 0.768 & 0.020 & 0.852 & 0.087 & 1.156 & 458 \\
% & w/o identity losses$^{\dagger}$
% & 0.353 & 0.245 & 0.629 & 0.057 & 1.638 & 293 \\
% & w/o motion losses
% & 0.795 & 0.011 & 0.901 & 0.090 & 1.171 & 548 \\

% \midrule
% \multirow{1}{*}{\textit{State-aware}}
% & w/o cold-start teacher
% & 0.721 & 0.028 & 0.834 & 0.079 & 1.188 & 406 \\

% \bottomrule
% \end{tabular}
% \end{table*}

% Main-paper ablations for Sec. 4.5 — ordered by method story
\begin{table*}[!b]
\centering
\caption{\textbf{Component ablations} on the 75-pair public cross-identity benchmark.
All variants continue from the same checkpoint for 5K steps. The gray row is the matched full recipe.
IDD: driver-identity similarity.
$^{\dagger}$Better motion scores may reflect driver copying with worse reference identity.}
\label{tab:ablation_components}

\small
\renewcommand{\arraystretch}{0.92}
\setlength{\tabcolsep}{4pt}

\begin{tabular}{llcccccc}
\toprule
& Variant
& ID$\uparrow$ & IDD$\downarrow$ & AED$\downarrow$ & APD$\downarrow$
& tLP$\downarrow$ & FVD$\downarrow$ \\
\midrule
\rowcolor{gray!20}
& Full recipe (continuation baseline)
& 0.707 & 0.023 & 0.816 & 0.071 & 1.254 & 389 \\

\specialrule{0.3pt}{1pt}{1pt}
\textit{Pixel cond.}
& Driver attention $w=0$$^{\dagger}$
& 0.585 & 0.210 & 0.679 & 0.067 & 1.690 & 273 \\

\specialrule{0.3pt}{1pt}{1pt}
\multirow{3}{*}{\textit{Cross-ID}}
& w/o cross-identity supervision
& 0.768 & 0.020 & 0.852 & 0.087 & 1.156 & 458 \\
& w/o identity losses$^{\dagger}$
& 0.353 & 0.245 & 0.629 & 0.057 & 1.638 & 293 \\
& w/o motion losses
& 0.795 & 0.011 & 0.901 & 0.090 & 1.171 & 548 \\

\specialrule{0.3pt}{1pt}{1pt}
\textit{State-aware}
& w/o cold-start teacher
& 0.721 & 0.028 & 0.834 & 0.079 & 1.188 & 406 \\

\bottomrule
\end{tabular}
\end{table*}
We ablate PixReenact's main components on the 75-pair public cross-identity
benchmark in Table~\ref{tab:ablation_components}. Each ablation continues the
same checkpoint for 5K steps, with the full recipe as reference, so absolute
values may differ from Table~\ref{tab:cross_benchmarks}. Qualitative effects are
shown in Appendix Figure~\ref{fig:ablation_qual}.

\noindent\textbf{Pixel-driven conditioning.} We ablate the driving-attention window of Sec.~\ref{subsec:pixel_driven}. Replacing our $\pm1$ window with only the aligned latent ($w=0$) hurts identity preservation: ID drops from 0.707 to 0.585, while driver-ID similarity (IDD) rises from 0.023 to 0.210. Meanwhile, AED, APD, and FVD improve, showing that better motion scores can reflect driver similarity rather than better reenactment. Local temporal context therefore better balances motion transfer and identity preservation.

\noindent\textbf{Cross-identity learning and corrective objectives.}
Removing cross-identity supervision worsens AED
($0.816\!\rightarrow\!0.852$), APD ($0.071\!\rightarrow\!0.087$), and FVD
($389\!\rightarrow\!458$), supporting its role in learning cross-identity motion transfer. Without identity losses, ID drops to 0.353 and IDD rises to 0.245 while
motion scores improve due to driver leakage. Without motion losses, ID rises to 0.795 but AED, APD, and FVD worsen. Both are needed to balance identity
preservation and motion transfer.

\noindent\textbf{State-aware distillation.}
Using the steady-state teacher for the first emission worsens AED
($0.816\!\rightarrow\!0.834$), APD ($0.071\!\rightarrow\!0.079$), and FVD
($389\!\rightarrow\!406$), despite slightly higher ID. This supports a dedicated
first-emission teacher when motion-aligned history is unavailable.

\noindent\textbf{Quality-latency trade-off.} Reducing 4-NFE to 3/2-NFE cuts latency from 238.9 to 211.4/186.6~ms but worsens motion and distribution quality. Appendix Table~\ref{tab:reduced_nfe} gives full PersonaLive comparisons.

Appendix Table~\ref{tab:ablation_additional} ablates DMD,
pseudo-target filtering, self-driving data, and reconstruction loss.
\section{Conclusion}
\label{sec:conclusion}

We introduce \textbf{PixReenact}, a pixel-conditioned framework for low-latency
streaming head-avatar reenactment. Across three cross-identity benchmarks and a
long-horizon streaming benchmark, PixReenact demonstrates robust cross-identity
reenactment under extreme viewpoints, occlusions, and pronounced facial
expressions, while maintaining reference identity over five-minute streams.
Its 4-NFE rolling formulation continuously emits four frames at 239~ms mean latency.

Challenges remain. Extreme viewpoints and severe occlusions still challenge
streaming reenactment, leaving room to improve motion fidelity and visual
quality. Cross-identity training uses costly pseudo targets from an offline
model, whose errors leave supervision imperfect despite corrective objectives.
Reducing this reliance is important future work. More broadly, our results
support direct visual conditioning with causal video generation for robust,
low-latency streaming head avatars.

\bibliography{iclr2027_conference}
\bibliographystyle{iclr2027_conference}

\appendix

\section{Implementation and Evaluation Details}
\label{sec:supp_details}

Unless stated otherwise, all details refer to the released \emph{wan-labels} model. The \emph{PL-labels} variant differs only in the pseudo-target generator, PersonaLive~\cite{personalive} instead of Wan-Animate~\cite{wan-animate}, and in the filtering thresholds noted below.

\subsection{Training Data Construction}
\label{subsec:supp_data}

\paragraph{Pseudo-pair construction.}
Each cross-identity training example pairs a reference image with a
4-second, $512^2$ driver clip of a different identity. For NeRSemble~\cite{nersemble}, VFHQ training set~\cite{vfhq}, and DH-FaceVid-1K~\cite{dhfacevid}, references and drivers are drawn from the corresponding training source. For FFHQ, we evenly sample 800 references from indices 00050-69999 and pair each with three drivers, yielding 2{,}400 candidate
pairs before filtering. The pseudo target is generated offline by Wan-Animate~\cite{wan-animate} (20 denoising steps) at
$512^2$. The PL-labels variant instead generates targets with PersonaLive~\cite{personalive} in its streaming mode on the same pairings.

\paragraph{Pseudo-target filtering.}
Pseudo targets are filtered for \emph{label quality} using identity, motion, and emotion consistency criteria: expression and pose agreement with the driver (AED $\leq 0.25$, APD $\leq 0.18$), emotion agreement (EmoSim $\geq 0.7$), and an
identity-routing margin
$\mathrm{ID}(\tilde{Y},I_{\mathrm{ref}})-\mathrm{ID}(\tilde{Y},D)\geq 0$
that rejects pseudo targets that copied the driver's identity. We never
filter on absolute identity similarity alone, which would remove mainly hard pairs in which the reference image and driver pose differ. PL labels exhibit a stronger driver-copy
failure mode, so their thresholds are stricter on emotion (EmoSim $\geq 0.84$) and margin ($\geq 0.1$), while the AED/APD thresholds are identical. Table~\ref{tab:supp_filtering} reports per-dataset counts.
Filtering metrics are computed with the training-time estimators
(Sec.~\ref{subsec:supp_losses}): DECA expression (50-D) and head-rotation
$\ell_1$ for AED/APD, EmoNet feature cosine for EmoSim, and ArcFace cosine
for the identity terms, each evaluated per frame on FAN face crops~\cite{FAN} and
averaged over the clip (full-frame resize when the detector misses a
frame). If no face is detected in the pseudo target at all, its identity
terms are undefined and the pair fails the margin gate, i.e., it is
discarded.

\begin{table}[h]
\centering
\caption{Pseudo-pair counts before (scored) and after quality filtering.}
\label{tab:supp_filtering}
\resizebox{0.6\textwidth}{!}{%
\begin{tabular}{lcccc}
\toprule
 & \multicolumn{2}{c}{wan-labels} & \multicolumn{2}{c}{PL-labels} \\
\cmidrule(lr){2-3}\cmidrule(lr){4-5}
Source & scored & kept & scored & kept \\
\midrule
NeRSemble       & 4{,}020  & 3{,}182  & 4{,}012  & 3{,}061 \\
VFHQ            & 1{,}450  & 1{,}106  & 1{,}447  & 1{,}084 \\
DH-FaceVid-1K   & 10{,}000 & 7{,}990  & 9{,}994  & 7{,}686 \\
FFHQ            & 2{,}400  & 1{,}283  & 2{,}400  & 1{,}642 \\
\midrule
Total           & 17{,}870 & 13{,}561 & 17{,}853 & 13{,}473 \\
\bottomrule
\end{tabular}}
\end{table}

\paragraph{Real self-driving clips.}
1{,}548 real 4-second clips (402 NeRSemble, 146 VFHQ, 1{,}000 DH-FaceVid-1K). The first frame is the reference image and the clip is both driver and ground truth. During student training, each step draws a self-driving sample with probability of $60\%$ and a cross-identity pseudo-pair otherwise.

\subsection{Model Architecture}
\label{subsec:supp_arch}
We use Wan2.1-T2V-1.3B~\cite{wan2.1} as the bidirectional video-diffusion backbone (30 transformer blocks, hidden dimension 1{,}536) and adopt CausVid's~\cite{causvid} causal generation framework. The model uses the Wan-VAE ($8\times$ spatial, $4\times$ temporal compression; one latent spans four frames), and all experiments run at $512^2$. The pretrained backbone weights and VAE are frozen. We train LoRA adapters (rank 64, $\alpha=64$) in the self-attention projections of all 30 blocks, together with the conditioning interface. Conditioning follows Sec.~\ref{subsec:pixel_driven}: one identity latent, window-aligned driving latents, and two context latents, with the restricted attention mask described there (driving window $w{=}1$). Text prompts are not used.

\subsection{Streaming Configuration}
\label{subsec:supp_streaming}
The students' rolling staircase uses four noise levels $t=[1000,750,500,250]$. Each forward pass admits one fresh noise latent at $t{=}1000$ (seed $42$), then advances all in-flight latents by one level, and emits the $t{=}250$ latent to $t{=}0$ (four frames). Context is the two most recent emitted latents. Conditioning keys/values (identity and driving tokens) are KV-cached across passes. Decoding of emitted latents uses the streaming VAE decoder with a persistent temporal cache. Streams start with a ramp state: the context slots hold causal-VAE latents of a short still clip of the reference image, and the first emission is aligned to driver frame $0$, so outputs are frame-aligned to the driver over the full stream. Driver frames are face-aligned $512^2$ crops encoded by the causal VAE. At inference time, encoding of the next driving chunk overlaps with denoising of the current chunk.

\subsection{Dense Score Teachers}
\label{subsec:supp_teachers}

\paragraph{Steady-state teacher.}
We train rank-$64$ LoRA adapters on the pretrained bidirectional
Wan2.1-T2V-1.3B backbone under the causal attention mask, using the deployed
rolling geometry (6-latent windows: 2 context + 4 video latents). The adapters
are trained for 27{,}500 steps from initialization (lr $2\times10^{-4}$),
followed by 10{,}000 adaptation steps (lr $10^{-4}$) on the filtered
pseudo-pairs with 10\% self-driving mixing. Batch size is $4$ per GPU on
$4$ NVIDIA H200 GPUs. Noise states mix 60\% staircase-patterned windows
($t=[0,0,250,500,750,1000]$) and 40\% dense timesteps
$t\sim\mathcal{U}(0.02,1.0)$, so the teacher covers both inference-like
mixed-noise states and arbitrary re-noise levels queried during distillation.
Both teachers retain the backbone's native flow-velocity parameterization but
are supervised with MSE on the implied clean-latent estimate
$\hat{z}^0=z^t-\sigma_t\hat{v}^t$ against the clean target $z^0$.

\paragraph{Cold-start teacher.}
Initialized from the steady-state teacher and fine-tuned for 3{,}500 steps (lr $10^{-4}$, batch 4) exclusively on first-emission states: windows whose context is the reference-initialized still clip, over cross pairs with large reference-driver pose gaps (plus 10\% self-driving transport, where the reference and context come from a still of the same person as the real driver clip). Timesteps emphasize high noise ($t\geq0.8$ with probability of $90\%$), matching the states at which this teacher is queried (first emission from high-noise).

\subsection{Student Distillation}
\label{subsec:supp_distill}

The student is trained through long self-rollouts spanning the full 4-second driver clip (30 latents, 120 frames), giving $\approx$28 consecutive emissions from a single cold start. Rollouts are always opened from the deployment cold-start state (reference-initialized context). Losses are applied to emitted latents only. For DMD, the emitted latent is re-noised to $t\sim\mathcal{U}(0.02,0.98)$ with high-noise emphasis ($t\geq0.6$ for steady-state emissions; $t\geq0.8$ with probability of $85\%$ for the first emission). The real score is the cold-start teacher at the first emission and the steady-state teacher afterwards. The DMD term is scaled $\times3$ on first-emission states. The online fake-score critic is initialized from the same checkpoint as the student and updated three times per generator update. The released model is obtained by successive DMD stages: an initial 30{,}000-update distillation from the dense teacher followed by shorter refinement stages of the same objective family. The final wan-labels stage (1{,}000 updates) uses exactly the configuration reported here.

\subsection{Training Objectives}
\label{subsec:supp_losses}

Table~\ref{tab:supp_losses} lists all loss terms, their weights, and the samples they apply to during student training. Pixel-space losses are computed on frames decoded jointly with the preceding latent (matching deployment decoding
statistics). \textbf{Training-time estimators:} ArcFace R100 (Glint360k)~\cite{arcface} for the identity and anti-leakage terms; DECA~\cite{deca} (FLAME~\cite{flame} coefficients) on FAN-cropped faces for expression and head pose~\cite{FAN}; EmoNet~\cite{emonet} (8-class, 128-D features)
for emotion. On first emission (cold-start) states, the identity/anti-leak/anchor group is scaled $\times1.5$ and the expression/emotion group $\times0.6$.

\begin{table}[h]
\centering
\caption{Student loss terms. \textit{self} refers to real self-driving samples, \textit{cross} to cross-identity pseudo-pairs.}
\label{tab:supp_losses}

\resizebox{0.85\textwidth}{!}{%
\begin{tabular}{llcl}
\toprule
Term & Definition & Weight & Applies to \\
\midrule
$\mathcal{L}_{\mathrm{DMD}}$ & surrogate of Eq.~\ref{eq:dmd} (main paper) & 1.0 ($\times3$ cold) & all \\
$\mathcal{L}_{\mathrm{anc}}$ & $\ell_2$ to target latent & 0.3 & all \\
$\mathcal{L}_{\mathrm{id}}$ & $\max(0,\,0.8-\cos(\phi(Y_k),\phi(I_{\mathrm{ref}})))$ & 1.0 & all$^{a}$ \\
$\mathcal{L}_{\mathrm{leak}}$ & $\max(0,\,\cos(\phi(Y_k),\phi(D_k))-0.25)$ & 1.0 & cross \\
$\mathcal{L}_{\mathrm{pose}}$ & $\ell_1$ DECA global rotation vs.\ driver & 1.2 & all \\
$\mathcal{L}_{\mathrm{expr}}$ & $\ell_1$ DECA expression+jaw vs.\ driver & 0.8 & all \\
$\mathcal{L}_{\mathrm{emo}}$ & $1-\cos$ EmoNet features vs.\ driver & 0.5 & all \\
$\mathcal{L}_{\mathrm{rec, LPIPS}}$ & vs.\ ground-truth frames & 1.0 & self \\
$\mathcal{L}_{\mathrm{rec, Adversarial}}$ & PatchGAN hinge (disc.\ lr $10^{-4}$) & 0.1 & self \\
\bottomrule
\end{tabular}%
}

\vspace{1pt}

\begin{minipage}{0.85\textwidth}
\footnotesize
\textsuperscript{a} For cross-identity samples, the identity anchor is the
reference image $I_{\mathrm{ref}}$, while for self-driving samples it is the aligned
real ground-truth frame.
\end{minipage}
\end{table}

\subsection{Optimization}
\label{subsec:supp_optim}

We train using the Adam optimizer with $\beta_1{=}0$, $\beta_2{=}0.999$, and weight decay of $10^{-4}$.
The generator and critic learning rates are $10^{-5}$ and the discriminator uses AdamW with learning rate $10^{-4}$. All learning rates are constant, with no warm-up or decay schedule and no gradient accumulation. The critic takes three updates per generator update, and the discriminator takes one update per generator step on which the adversarial loss is active (real self-driving steps). No global training seed is fixed, while all inference and evaluation use seed 42. The trainable parameters are the LoRA adapters and conditioning interface, 17.7M parameters in total ($\approx$1.3\% of the 1.3B backbone). The final student stage uses batch size $1$ per GPU on 4 NVIDIA H200 GPUs, bfloat16 mixed precision, and gradient checkpointing.

\subsection{Runtime Measurement Protocol}
\label{subsec:supp_runtime}

\paragraph{Runtime measurement protocol.}
All runtime measurements use a single NVIDIA H200 GPU. Per-stage latencies
(Tab.~\ref{tab:latency_analysis}) are measured on the first 10 drivers of the
public cross-reenactment benchmark. Before timing, we perform one untimed
warm-up invocation to stabilize GPU clocks and caches. Each PixReenact clip is
then measured in a fresh process. Therefore, we additionally exclude the startup
ramp and first steady-state chunk, which absorbs one-time kernel compilation
costs. For each video, we report the median latency over the remaining
steady-state chunks and aggregate these values as mean $\pm$ std across the
10 videos.

Emission latency measures the time from admitting a chunk's noise latent until
its four output frames are decoded. Throughput is the corresponding steady-state
rate, computed as $\mathrm{FPS}=4/\mathrm{latency}$. Driver-VAE encoding is
excluded because, in streaming use, encoding of the next driving chunk overlaps
with denoising of the current chunk. Its cost is reported separately in
Tab.~\ref{tab:latency_analysis}. PersonaLive is evaluated under the same
steady-state timing definition using its official streaming pipeline.

\begin{table*}[t]
\centering
\caption{\textbf{Per-stage runtime of the two streaming pipelines}
(per emitted 4-frame chunk, $512^2$, steady state, single H200;
mean $\pm$ std over 10 drivers). FPS is computed as $4/$total emission
latency. Startup and one-time kernel warm-up are excluded.
Driver VAE encode (\textit{italic}) is excluded from our emission latency:
benchmark drivers enter as precomputed latents, and in live use the next
chunk's encode overlaps the current chunk's denoising. PersonaLive's
reference-UNet refresh fires at most three times per stream and is amortized.}
\label{tab:latency_analysis}
\begin{tabular}{llr}
\toprule
Pipeline & Stage & Lat.\,$\downarrow$ \\
\midrule
\multirow{8}{*}{PixReenact}
 & Window assembly (ctx + cond concat)            & 0.1\std{0.0} \\
 & Denoising DiT forward (4-level staircase, 1 pass) & 155.1\std{0.2} \\
 & Flow$\to x_0$ + re-noise (scheduler)           & 0.7\std{0.1} \\
 & Streaming VAE decode (4 frames, cached)        & 82.9\std{0.5} \\
 & Buffer bookkeeping (shift context/buffers)     & 0.1\std{0.0} \\
 & \textbf{Total emission latency}                & \textbf{238.9}\std{0.7} \\
 & \textbf{Generation throughput (FPS)}           & \textbf{16.74} \\
 & \textit{Driver VAE encode (streaming, amortized)} & \textit{45.5}\std{0.0} \\
\midrule
\multirow{11}{*}{PersonaLive}
 & Keypoint extraction                            & 5.3\std{0.1} \\
 & Face crop (4 frames)                           & 0.3\std{0.0} \\
 & Motion encoder                                 & 6.0\std{0.1} \\
 & Pose guider                                    & 1.1\std{0.0} \\
 & Denoising UNet (16-frame window, 1 step)       & 133.6\std{3.8} \\
 & Scheduler step                                 & 0.3\std{0.0} \\
 & Reference-UNet keyframe refresh (amortized)    & 0.1\std{0.1} \\
 & VAE decode (4 frames)                          & 61.2\std{0.0} \\
 & Host ops (resize, kps render, GPU$\to$CPU)     & 2.8\std{0.0} \\
 & \textbf{Total emission latency}                & \textbf{211.1}\std{3.6} \\
 & \textbf{Generation throughput (FPS)}           & \textbf{18.95} \\
\bottomrule
\end{tabular}
\end{table*}

\noindent\textbf{Quality-latency trade-off.}
We additionally vary the number of denoising steps while keeping all other
inference settings fixed. Table~\ref{tab:reduced_nfe} compares these schedules
with PersonaLive under the same runtime protocol on the 68-pair
extreme-condition benchmark. Reducing PixReenact from 4-NFE to 3-NFE and
2-NFE lowers latency from 238.9~ms to 211.4~ms and 186.6~ms, respectively,
but generally degrades motion and distribution quality. Notably, 3-NFE has
nearly identical latency to PersonaLive (211.4 vs.\ 211.1~ms), while the
4-NFE setting provides the strongest overall PixReenact quality. We therefore
use 4-NFE throughout the main experiments.
\begin{table*}[t]
\centering
\caption{\textbf{Quality-latency trade-off.}
Quality metrics are measured on the 68-pair extreme-condition benchmark,
while latency and FPS use the first 10 public-benchmark drivers under the same
runtime protocol for all methods. Fewer NFEs reduce latency but generally
degrade motion and distribution quality. Best and second-best results are shown
in \textbf{bold} and \underline{underline}, respectively.}
\label{tab:reduced_nfe}
\small
\begin{tabular}{lccccccc}
\toprule
Method / NFE
& ID$\uparrow$
& AED$\downarrow$
& APD$\downarrow$
& tLP$\downarrow$
& FVD$\downarrow$
& Lat.\,$\downarrow$
& FPS$\uparrow$ \\
\midrule

PersonaLive
& 0.472
& 1.295
& \textbf{0.135}
& \underline{1.866}
& \textbf{643.5}
& \underline{211.1}
& \underline{18.95} \\

\midrule

PixReenact, 4 (1000,750,500,250)
& 0.727
& \textbf{1.135}
& \underline{0.169}
& \textbf{1.853}
& \underline{662.5}
& 238.9
& 16.74 \\

PixReenact, 3 (1000,500,250)
& \underline{0.731}
& \underline{1.145}
& 0.174
& 2.008
& 682.9
& 211.4
& 18.92 \\

PixReenact, 2 (1000,250)
& \textbf{0.770}
& 1.191
& 0.187
& 1.986
& 753.6
& \textbf{186.6}
& \textbf{21.44} \\

\bottomrule
\end{tabular}
\end{table*}

\subsection{Benchmark Construction}
\label{subsec:supp_benchmarks}

\paragraph{Public cross-reenactment (75 pairs, Table~\ref{tab:cross_benchmarks}).}
Drivers are the 75 self-reenactment clips in fixed order (clips 1-25 = TalkingHead-1KH, 26-75 = VFHQ-Test). The reference image for pair $i$ is FFHQ image $(i-1) \bmod 50$, so each of the first 50 FFHQ images appears once or
twice.

\paragraph{NeRSemble (75 total pairs: 70 cross + 5 self, Tables~\ref{tab:cross_benchmarks} and~\ref{tab:nersemble_self}).}
The test pool contains 15 held-out subjects (NeRSemble IDs 82, 89, 94, 99, 100, 133, 145, 186, 371, 383, 388, 422, 443, 445, 475), identity-disjoint from training. Five subjects (388, 422, 443, 445, 475) provide the reference stills, and each reference is driven by a clip of every one of the 15 subjects: the five driver$=$reference combinations form the self-reenactment set and the remaining 70 the cross set. Clips are center-square cropped, temporally resampled to 30\,fps, and evaluated at $512^2$.

\paragraph{Extreme-condition (68 pairs, Table~\ref{tab:extreme_benchmark}).}
Candidate CelebV-HQ clips are ranked using the dataset's attribute annotations, scoring appearance attributes that occlude the face (mask, sunglasses, hat) and actions that produce large or fast head motion or hand/object occlusion (turn, look around, head wagging, head shake, nod, drink, eat, smoke, play instrument, make a face). Ranked candidates are then verified automatically (a head-pose probe records the median $|$yaw$|$/$|$pitch$|$ over each clip's hardest span, and a MediaPipe FaceMesh scan records per-frame detection misses) and reviewed manually. The final 68 drivers are grouped into four groups: extreme yaw (25), extreme pitch (17), accessory occlusion (16), and hand occlusion (10). References are FFHQ images assigned in the public cross-reenactment benchmark order, and drivers are center-square cropped and resized to $512^2$. Table~\ref{tab:extreme_clips} lists all 68 driver clips.
\begin{table*}[t]
\centering
\caption{\textbf{Driver clips of the extreme-condition benchmark.} The 68
CelebV-HQ clip identifiers by group.}
\label{tab:extreme_clips}
\footnotesize
\begin{tabular}{lp{0.7\textwidth}}
\toprule
Group & CelebV-HQ clip IDs \\
\midrule
Extreme yaw (25) & \texttt{-3xE76Q0yLs\_6\_3, G4f\_opQO7MA\_4\_0, I66kXy4rvME\_3, Lrpy2CCnJR0\_17, NkjnK6aSrqM\_4, QgedeMGTh5E\_9, RyevycwXzn4\_7, SnBObwk5JF0\_37, VCFAT8Hj7Qo\_0\_0, WGvhuhHYhiA\_14, YXetIu8JOzI\_3, cWTK6ttZUw0\_7, dhSpE\_2OF40\_9\_0, ecsatgCN6Y8\_1, hGsR1fkUJqY\_10, hOB4Qm1IiOY\_0, i1UEj\_6T1RE\_1, iViIy5ci5yU\_24, mpVMM7eqTKs\_8\_0, sb6b4r8no1w\_14\_1, t7BPFTnDsJc\_1, x99FBmR6OrI\_0\_0, xGoRE2NsRGI\_9, yprYw3FQUpQ\_0, zW2T9TkSIcM\_5} \\[2pt]
Extreme pitch (17) & \texttt{-kFJKQvCrkY\_1, 7NHQrVt1yOU\_2, 8w0b64kOPss\_0, 9DTvF8iBli8\_1\_3, AcvWJ8bgA8w\_3, Fn40RI8ving\_1, Jipg0KQ4id0\_3, N4XB-3O7UTk\_0\_0, PFBrK7pbqCU\_18, RLbry-3z8yQ\_2, Yg6sZ2htZfc\_1, dildzih2RPI\_3\_0, jeycN3BAkt8\_38\_0, mLh4ZKG50mk\_0\_0, od6IxUWPMcs\_0, qAHkYcmcFfc\_6, tnxXQgmHwrk\_2} \\[2pt]
Accessory occlusion (16) & \texttt{7FyjCUDR0IM\_17, CPKMurScLSE\_15\_0, Fn40RI8ving\_7, HK6TmhLEd2E\_12, LjJ276Dh1jA\_44, Tn0Ge0GIoGA\_10, U8PnHiRRg-s\_6, XDmfBTgIRUg\_12, XDmfBTgIRUg\_15, \_wiIxAcoTF4\_4, i4ivf-ZmorE\_28, qs7FL5DVDMA\_10, unmu4yKfBg0\_9, xuphB-zbmPk\_26, yLElO7gxXPo\_43, z7VS7DVHTbc\_1\_0} \\[2pt]
Hand occlusion (10) & \texttt{2Ke4u7vbLE0\_0, M9FAInQwCm0\_3, NgtK-eesI7g\_0, PP9l4LP0WPI\_4, Q9B0pFDCviw\_102, dfrJhivMJJY\_0, hSiFAK-p-Lg\_20, v60fsgGSoDU\_0, xu\_UgfGriXA\_3, yth5JoZKnGQ\_1} \\
\bottomrule
\end{tabular}
\end{table*}

\paragraph{Long-horizon streaming (28 pairs, Figure~\ref{fig:id_sim_long_videos}).}
To evaluate identity stability over long streams, we construct a benchmark of
28 continuous 5-minute cross-identity clips. The drivers comprise 10 long-form
Barack Obama videos collected from YouTube, 15 videos from the WDA subset of
HDTF~\cite{hdtf}, and 3 long clips extracted from the raw TalkingHead-1KH
validation videos~\cite{talkinghead1kh}. Each driver is paired with a fixed FFHQ
reference image, following the reference order of the public cross-reenactment
benchmark, with driver and reference identities always different. Each method
processes the full 5-minute sequence continuously, without resetting its
temporal state. We measure AdaFace similarity to the reference throughout the
stream and aggregate the results over five one-minute intervals. For each clip,
scores are first averaged within each minute and then averaged across clips.

\subsection{Baseline Implementation Details}
\label{subsec:supp_baselines}

All baselines run from their official repositories, released checkpoints,
and default configurations, with their own native preprocessing. We list
only the deviations required for full-length, frame-aligned evaluation.
\textbf{LivePortrait}: official release; per-frame animation with its
whole-clip tracking pass; no modifications.
\textbf{Wan-Animate}: official 14B release at 20 denoising steps; native
77-frame windows with the repository's sliding-window continuation for
full-length clips.
\textbf{Wan-Animate-2-Base}: official release (37 frames per
pass); full-length outputs are produced by sequential client-side chunking,
as the release provides no long-video continuation.
\textbf{X-NeMo}: official release with its native long-video context
scheduler (24-frame windows, overlap 4); evaluated in its native crop
space against identically cropped references.
\textbf{PersonaLive}: official streaming (online) pipeline, unmodified.
\textbf{X-Portrait}: official checkpoint; the required per-pair
best-frame anchor is selected automatically by matching driver head pose
to the reference via FAN landmarks.
\textbf{MegActor-$\Sigma$}: official weights and flags (25 steps, guidance
2); the default 2.5-second generation limit is raised to the full clip
length.
\textbf{FlashPortrait}: official checkpoint on Wan2.1-I2V-14B with a fixed
generic text prompt for all clips; its scheduler requires $\geq$51 latents,
so shorter clips cap the context accordingly, and the official output
drops the first frame and truncates to $4k{+}1$ frames, which we restore
by duplicating $\leq$3 boundary frames.
For every method, outputs are fitted to exactly the driver frame count by
frame-indexed rewriting when the difference is $\leq$3 frames (float-fps
container rounding); larger mismatches abort the sample rather than
re-time the video.

\subsection{Evaluation Protocol}
\label{subsec:supp_eval}
All methods are scored on the same evaluation clips at $512^2$, full length, with outputs frame-aligned to the driver (frame $i$ of every method corresponds to driver frame $i$). \textbf{ID / IDD:} cosine similarity of AdaFace~\cite{adaface} (CVLface IR-101, WebFace12M) embeddings between generated frames and the reference image (ID) or the corresponding driver frame (IDD). Faces are detected, aligned and cropped to $112^2$, as in the standard AdaFace pipeline. For speed, we do not embed every frame of the clip, but use ${\sim}60$ uniformly sampled frames.
\textbf{AED / APD:} SMIRK~\cite{smirk} codes on face crops (FAN bounding
box, DECA-convention crop, $224^2$) computed for every frame. AED is the $\ell_1$ distance over expression, eyelid, and jaw parameters
(55-D) and APD over the 3 head-pose parameters, between generated and
driver frames.
\textbf{tLP:}~\cite{tLP} mean over consecutive-frame pairs of
$|d(Y_t,Y_{t+1})-d(R_t,R_{t+1})|\times100$, where $d$ is
LPIPS~\cite{lpips} (AlexNet) and $R$ is the reference sequence (ground
truth for self-reenactment, driver for cross).
\textbf{FVD:}~\cite{FVD} set-level with an I3D backbone, where each clip
contributes up to eight 16-frame windows (stride 16) spread uniformly over
the clip, resized to $224^2$. Real statistics come from the ground-truth
(self) or driver (cross) clips.
\textbf{PSNR / SSIM / LPIPS / $\ell_1$:} full-frame at $512^2$ against
ground truth, self-reenactment only.
X-NeMo~\cite{x-nemo} is evaluated in its native crop space against identically cropped
references. Quality tables report the per-clip mean (\,$\pm$ std where shown), while FVD is a single set-level value.

\subsection{Limitations of Evaluation Metrics}
\label{sec:metric_discussion}

The metrics used in the main paper capture complementary aspects of reenactment quality, but no individual metric fully measures successful cross-identity reenactment. In particular, identity preservation and motion transfer must be considered jointly: reproducing the driver's face more directly can improve agreement with the driving motion while failing to preserve the reference identity.

AED and APD measure distances between expression and pose parameters estimated by SMIRK~\cite{smirk}, rather than distances to ground-truth facial motion. This distinction is especially relevant for our extreme-condition benchmark in Table~\ref{tab:extreme_benchmark}. SMIRK reports sensitivity to occlusions and shows degraded reconstructions under occlusions and extreme head poses. Consequently, APD and AED estimates may become less reliable precisely on some of the most challenging frames in this benchmark. FVD~\cite{FVD} also has known limitations as a measure of temporal video quality. Prior work has shown that FVD can be strongly influenced by individual-frame appearance and can be relatively insensitive to temporal corruptions~\cite{fvd_content_bias}. Thus, a lower FVD does not necessarily imply better motion transfer or temporal realism in isolation.

Our component ablations provide an additional example specific to
cross-identity reenactment. Removing the identity losses, or restricting the
driving attention to only the aligned latent, increases driver-identity
leakage while improving several motion or distribution metrics
(Table~\ref{tab:ablation_components}). These improvements do not reflect better cross-identity reenactment: the generated
video becomes more similar to the driver at the expense of preserving the
reference identity. We therefore interpret ID, AED/APD, tLP, and FVD jointly
and complement the quantitative results with qualitative comparisons.

\section{Additional Results}
\label{sec:supp_additional_results}

\begin{figure*}[t]
    \centering
    \includegraphics[width=\textwidth]{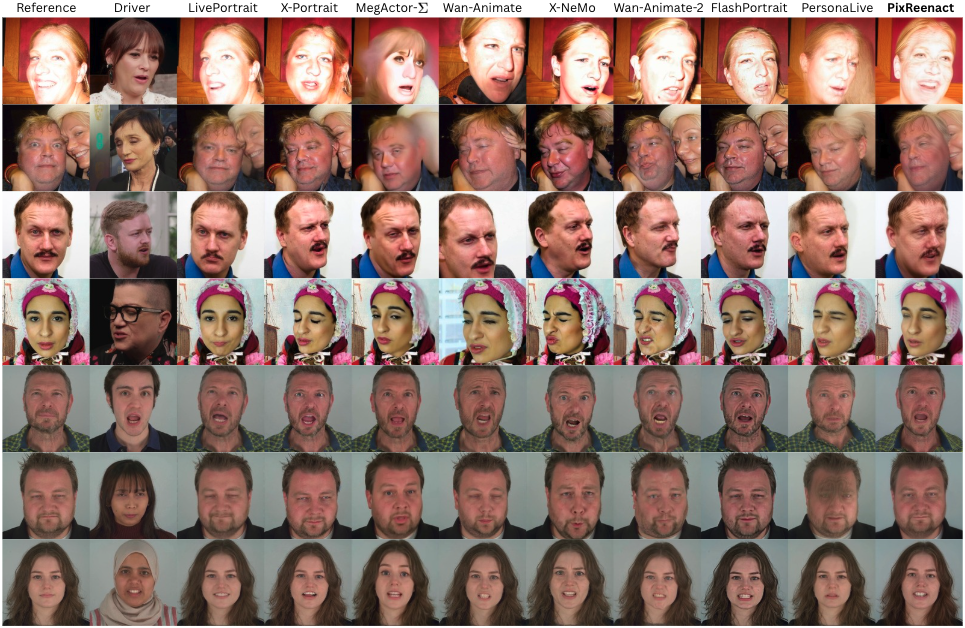}
    \caption{\textbf{Additional cross-identity reenactment examples.}
    Qualitative comparisons on the public (rows 1-4) and NeRSemble
    (rows 5-7) benchmarks against all evaluated baselines. PixReenact
    preserves the reference appearance while transferring the driver's
    facial motion.}
    \label{fig:qual_cross_additional}
\end{figure*}

\paragraph{Cross-identity reenactment.}
Figure~\ref{fig:qual_cross_additional} provides additional qualitative
comparisons on the public and NeRSemble benchmarks. Across varied identities, poses, and expressions, PixReenact preserves the reference appearance while following the driver's facial motion. The figure also compares against all
per-frame, offline, and streaming baselines evaluated in the main paper.

\paragraph{Sensitivity to the pseudo-target generator.}
The main paper reports the \textit{wan-labels} model as PixReenact. To test whether the
method depends strongly on the particular model used to construct the
cross-identity pseudo targets, Table~\ref{tab:pseudo_target_generator} compares
it with an otherwise equivalent model trained using PersonaLive-generated
pseudo targets. Neither source consistently dominates across benchmarks or
metrics: \textit{PL-labels} performs slightly better on several motion and temporal
metrics, while \textit{wan-labels} provides stronger identity preservation on the public
and NeRSemble benchmarks and lower FVD on the public benchmark. The two models
also perform nearly identically on the extreme-condition benchmark. Overall,
the results indicate that the proposed training framework is not specific to a
single pseudo-target generator.

\begin{table*}[t]
\centering
\caption{\textbf{Sensitivity to the pseudo-target generator.}
We compare PixReenact trained with Wan-Animate pseudo targets (wan-labels)
against the same framework trained with PersonaLive pseudo targets (PL-labels) on cross-reenactment benchmarks.
Best result within each benchmark is shown in \textbf{bold}.}
\label{tab:pseudo_target_generator}
\setlength{\tabcolsep}{5pt}
\begin{tabular}{llccccc}
\toprule
Benchmark & Pseudo targets
& ID$\uparrow$ & AED$\downarrow$ & APD$\downarrow$
& tLP$\downarrow$ & FVD$\downarrow$ \\
\midrule

\multirow{2}{*}{Public (75)}
& Wan-Animate
& \textbf{0.709} & 0.790 & \textbf{0.066} & 1.234 & \textbf{377.1} \\
& PersonaLive
& 0.704 & \textbf{0.777} & 0.070 & \textbf{1.126} & 402.9 \\
\midrule

\multirow{2}{*}{NeRSemble (70)}
& Wan-Animate
& \textbf{0.753} & 0.714 & 0.0426 & 0.803 & 264.8 \\
& PersonaLive
& 0.735 & \textbf{0.689} & \textbf{0.0386}
& \textbf{0.786} & \textbf{250.8} \\
\midrule

\multirow{2}{*}{Extreme (68)}
& Wan-Animate
& 0.727 & \textbf{1.135} & \textbf{0.169}
& \textbf{1.853} & 662.5 \\
& PersonaLive
& \textbf{0.729} & 1.139 & 0.175
& 1.876 & \textbf{653.7} \\
\bottomrule
\end{tabular}
\end{table*}

\begin{figure*}[t]
    \centering
    \includegraphics[width=\textwidth]{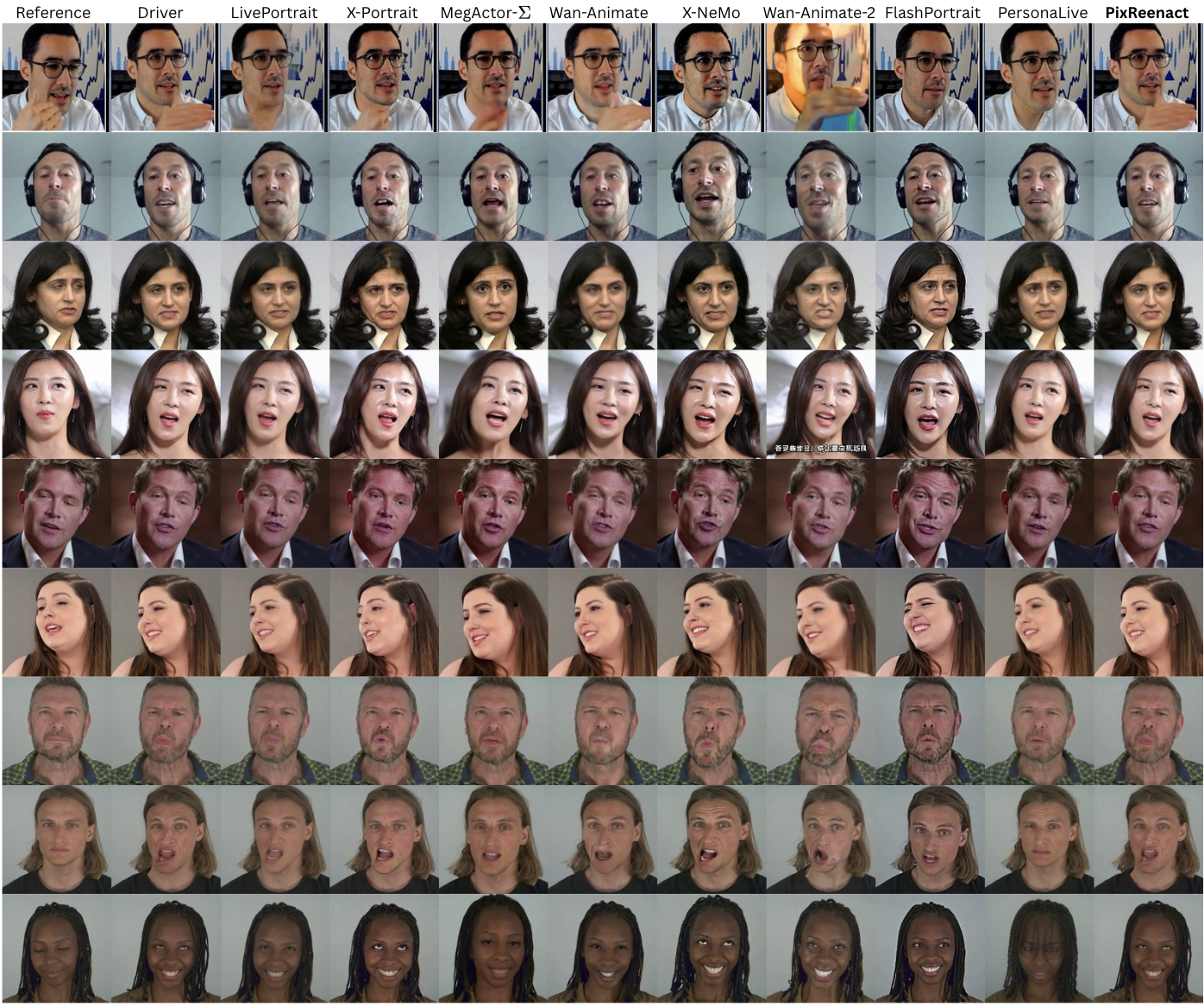}
    \caption{\textbf{Qualitative self-reenactment comparisons.}
    Rows 1-3 show TalkingHead-1KH, rows 4-6 VFHQ-Test, and rows 7-9
    NeRSemble. PixReenact closely reconstructs the driving motion and appearance
    while maintaining high visual fidelity across all three datasets.}
    \label{fig:qual_self}
\end{figure*}

\paragraph{Self-reenactment.}
Tables~\ref{tab:public_benchmark_self} and~\ref{tab:nersemble_self} report
self-reenactment results on TalkingHead-1KH, VFHQ-Test, and a small 5-sample
NeRSemble benchmark, with qualitative comparisons shown in
Figure~\ref{fig:qual_self}. In this setting, the driving video has the same
identity as the reference and therefore provides a strong appearance cue.
PixReenact achieves reconstruction quality close to the VAE reconstruction
baseline, obtained by directly encoding and decoding the ground-truth video
with the frozen Wan-VAE. Importantly, its strong cross-identity performance
shows that this does not arise simply from copying the driver: the same model
transfers motion from a different identity while preserving the reference
appearance.

\begin{table*}[t]
\centering
\caption{\textbf{Self-reenactment on the public benchmark} (75 clips:
TalkingHead-1KH + VFHQ-Test). All methods are re-run at $512^2$ on full clips
at native FPS with frame-aligned evaluation; PersonaLive uses streaming mode,
and X-NeMo is evaluated in its native crop space. Offline methods have no
emission latency. \emph{VAE round-trip} gives the reconstruction ceiling and
is excluded from comparison. For streaming methods, the better result between
PersonaLive and PixReenact is shown in \textbf{bold}.}
\label{tab:public_benchmark_self}
\setlength{\tabcolsep}{2.2pt}
\resizebox{\textwidth}{!}{%
\begin{tabular}{llccccccccccc}
\toprule
& Method
& PSNR$\uparrow$ & SSIM$\uparrow$ & LPIPS$\downarrow$ & L1$\downarrow$
& ID$\uparrow$ & AED$\downarrow$ & APD$\downarrow$
& tLP$\downarrow$ & FVD$\downarrow$ & Lat.\,$\downarrow$ & FPS$\uparrow$ \\
\midrule

\multicolumn{13}{c}{\textit{TalkingHead-1KH (25 clips)~\cite{talkinghead1kh}}} \\
\midrule

& \textit{VAE round-trip}
& \textit{38.25} & \textit{0.971} & \textit{0.015} & \textit{0.009}
& \textit{0.857} & \textit{0.084} & \textit{0.0019}
& \textit{0.184} & \textit{5.7} & -- & -- \\

\midrule
\multirow{2}{*}{\textit{Per-frame}}
& LivePortrait
& 19.90 & 0.717 & 0.238 & 0.069 & 0.881 & 0.371 & 0.0256
& 0.856 & 154.2 & -- & 18.90 \\
& X-Portrait
& 20.32 & 0.722 & 0.203 & 0.063 & 0.839 & 0.366 & 0.0226
& 0.832 & 209.1 & -- & 0.65 \\

\midrule
\multirow{5}{*}{\textit{Offline}}
& MegActor-$\Sigma$
& 17.47 & 0.664 & 0.284 & 0.086 & 0.825 & 0.460 & 0.0471
& 1.138 & 119.0 & -- & 0.49 \\
& Wan-Animate
& 21.22 & 0.750 & 0.184 & 0.058 & 0.763 & 0.393 & 0.0159
& 1.530 & 106.4 & -- & 0.41 \\
& X-NeMo
& 17.40 & 0.628 & 0.294 & 0.088 & 0.816 & 0.387 & 0.0291
& 1.067 & 189.2 & -- & 1.37 \\
& Wan-Animate-2
& 16.96 & 0.634 & 0.338 & 0.107 & 0.655 & 0.600 & 0.0462
& 3.990 & 381.7 & -- & 0.29 \\
& FlashPortrait
& 17.19 & 0.646 & 0.287 & 0.095 & 0.839 & 0.476 & 0.0566
& 1.621 & 203.4 & -- & 3.03 \\

\midrule
\multirow{2}{*}{\textit{Streaming}}
& PersonaLive
& 19.55 & 0.696 & 0.222 & 0.070 & \textbf{0.869}
& 0.363 & 0.0210 & 0.938 & 199.4
& \textbf{211.1} & \textbf{18.95} \\
& PixReenact
& \textbf{37.48} & \textbf{0.965} & \textbf{0.019} & \textbf{0.009}
& 0.855 & \textbf{0.096} & \textbf{0.0028}
& \textbf{0.237} & \textbf{8.2} & 238.9 & 16.74 \\

\midrule
\multicolumn{13}{c}{\textit{VFHQ-Test (50 clips)~\cite{vfhq}}} \\
\midrule

& \textit{VAE round-trip}
& \textit{35.73} & \textit{0.950} & \textit{0.020} & \textit{0.011}
& \textit{0.870} & \textit{0.090} & \textit{0.0022}
& \textit{0.243} & \textit{1.8} & -- & -- \\

\midrule
\multirow{2}{*}{\textit{Per-frame}}
& LivePortrait
& 18.97 & 0.639 & 0.284 & 0.075 & 0.889 & 0.318 & 0.0218
& 1.481 & 95.1 & -- & 18.90 \\
& X-Portrait
& 19.43 & 0.660 & 0.221 & 0.065 & 0.855 & 0.330 & 0.0270
& 0.852 & 112.7 & -- & 0.65 \\

\midrule
\multirow{5}{*}{\textit{Offline}}
& MegActor-$\Sigma$
& 17.15 & 0.581 & 0.341 & 0.093 & 0.793 & 0.430 & 0.0449
& 1.084 & 111.4 & -- & 0.49 \\
& Wan-Animate
& 21.45 & 0.710 & 0.174 & 0.051 & 0.825 & 0.311 & 0.0139
& 0.997 & 57.1 & -- & 0.41 \\
& X-NeMo
& 17.09 & 0.567 & 0.318 & 0.092 & 0.839 & 0.339 & 0.0255
& 1.264 & 100.5 & -- & 1.37 \\
& Wan-Animate-2
& 18.49 & 0.618 & 0.267 & 0.074 & 0.780 & 0.397 & 0.0299
& 2.348 & 122.9 & -- & 0.29 \\
& FlashPortrait
& 17.32 & 0.587 & 0.313 & 0.089 & 0.844 & 0.395 & 0.0446
& 1.655 & 115.2 & -- & 3.03 \\

\midrule
\multirow{2}{*}{\textit{Streaming}}
& PersonaLive
& 19.04 & 0.631 & 0.248 & 0.070 & \textbf{0.878}
& 0.332 & 0.0212 & 1.317 & 83.8
& \textbf{211.1} & \textbf{18.95} \\
& PixReenact
& \textbf{34.03} & \textbf{0.930} & \textbf{0.034} & \textbf{0.014}
& 0.860 & \textbf{0.117} & \textbf{0.0045}
& \textbf{0.377} & \textbf{5.5} & 238.9 & 16.74 \\

\bottomrule
\end{tabular}}
\end{table*}
\begin{table*}[t]
\centering
\caption{\textbf{Self-reenactment on the NeRSemble~\cite{nersemble} test set}
(5 clips of held-out subjects). \emph{VAE round-trip} marks the shared-VAE
reconstruction ceiling and is excluded from comparison. For streaming methods,
the better result is shown in \textbf{bold}.}
\label{tab:nersemble_self}
\resizebox{\textwidth}{!}{%
\begin{tabular}{llccccccccc}
\toprule
& Method & PSNR$\uparrow$ & SSIM$\uparrow$ & LPIPS$\downarrow$ & L1$\downarrow$
& ID$\uparrow$ & AED$\downarrow$ & APD$\downarrow$ & tLP$\downarrow$ & FVD$\downarrow$ \\
\midrule

& \textit{VAE round-trip}
& \textit{38.19} & \textit{0.959} & \textit{0.023} & \textit{0.007}
& \textit{0.815} & \textit{0.111} & \textit{0.0024} & \textit{0.259} & \textit{8.8} \\

\midrule
\multirow{2}{*}{\textit{Per-frame}}
& LivePortrait
& 25.88 & 0.810 & 0.168 & 0.030 & 0.848 & 0.341 & 0.0105 & 0.943 & 168.4 \\
& X-Portrait
& 24.31 & 0.739 & 0.157 & 0.035 & 0.822 & 0.430 & 0.0168 & 0.841 & 139.1 \\

\midrule
\multirow{5}{*}{\textit{Offline}}
& MegActor-$\Sigma$
& 20.07 & 0.704 & 0.279 & 0.062 & 0.743 & 0.642 & 0.0297 & 1.190 & 319.3 \\
& Wan-Animate
& 24.99 & 0.797 & 0.161 & 0.033 & 0.752 & 0.510 & 0.0151 & 1.679 & 209.3 \\
& X-NeMo
& 23.36 & 0.765 & 0.186 & 0.043 & 0.784 & 0.368 & 0.0136 & 0.842 & 133.8 \\
& Wan-Animate-2
& 23.10 & 0.767 & 0.212 & 0.042 & 0.667 & 0.508 & 0.0222 & 2.181 & 176.0 \\
& FlashPortrait
& 21.30 & 0.695 & 0.234 & 0.054 & 0.811 & 0.538 & 0.0330 & 1.794 & 318.3 \\

\midrule
\multirow{2}{*}{\textit{Streaming}}
& PersonaLive
& 23.95 & 0.770 & 0.200 & 0.042 & 0.794 & 0.697 & 0.0223 & 1.219 & 251.7 \\
& PixReenact
& \textbf{37.53} & \textbf{0.949} & \textbf{0.031} & \textbf{0.009}
& \textbf{0.814} & \textbf{0.129} & \textbf{0.0033} & \textbf{0.302}
& \textbf{13.9} \\

\bottomrule
\end{tabular}}
\end{table*}

\begin{figure*}[t]
    \centering
    \includegraphics[width=0.8\textwidth]{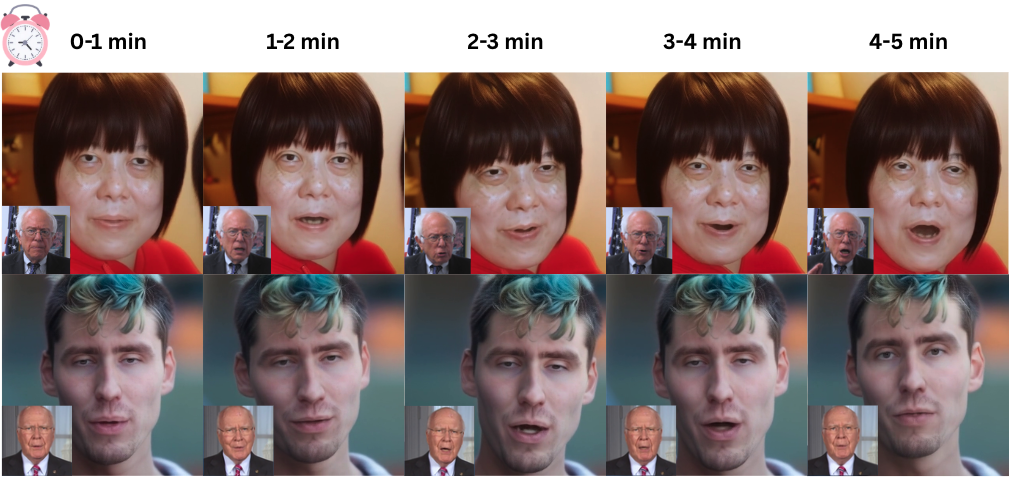}
    \caption{\textbf{Long-horizon streaming examples.} PixReenact outputs from two five-minute cross-identity videos, shown at one-minute intervals. The small images show the corresponding driver frames. PixReenact keeps a stable appearance while following the driver over time.}
    \label{fig:long_horizon_qual}
\end{figure*}

\paragraph{Long-horizon streaming.}
Figure~\ref{fig:long_horizon_qual} shows PixReenact outputs sampled at
one-minute intervals from two five-minute cross-identity streams. The generated
appearance remains stable throughout the sequence while following the changing
driver motion, complementing the quantitative identity-stability analysis in
Figure~\ref{fig:id_sim_long_videos}.

\paragraph{Single-window cross-reenactment.}
To separate cross-reenactment quality from the difficulty of long-video
continuation, Table~\ref{tab:single_window_evaluation} evaluates the public
benchmark using only the first 77 frames of each driver, corresponding to a
single Wan-Animate~\cite{wan-animate} generation window. The overall trends remain similar to the
full-length evaluation: PixReenact maintains strong identity preservation and
video quality, while the relative differences in motion metrics remain. This
shows that the main cross-reenactment results are not explained solely by the
need to stitch or continue long clips.
\begin{table*}[t]
\centering
\caption{\textbf{Single-window cross-reenactment.} We use the same 75
reference-driver pairs as the public benchmark (Table~\ref{tab:cross_benchmarks}),
but trim each driver to its first 77 frames ($\approx$3\,s), corresponding to
one Wan-Animate generation window. Thus, offline baselines require no long-video
continuation or stitching. Better streaming results are shown in bold. FVD is set-level.}
\label{tab:single_window_evaluation}
\begin{tabular}{llccccc}
\toprule
& Method & ID$\uparrow$ & AED$\downarrow$ & APD$\downarrow$
& tLP$\downarrow$ & FVD$\downarrow$ \\
\midrule

\multirow{2}{*}{\textit{Per-frame}}
& LivePortrait
& 0.892 & 1.075 & 0.207 & 1.151 & 636.6 \\
& X-Portrait
& 0.716 & 0.911 & 0.106 & 1.546 & 639.5 \\

\midrule
\multirow{5}{*}{\textit{Offline}}
& MegActor-$\Sigma$
& 0.570 & 0.829 & 0.100 & 1.911 & 620.0 \\
& Wan-Animate
& 0.586 & 0.935 & 0.085 & 1.609 & 690.0 \\
& X-NeMo
& 0.619 & 0.765 & 0.064 & 1.373 & 510.8 \\
& Wan-Animate-2
& 0.655 & 0.865 & 0.090 & 3.633 & 690.7 \\
& FlashPortrait
& 0.775 & 0.858 & 0.107 & 3.697 & 715.9 \\

\midrule
\multirow{2}{*}{\textit{Streaming}}
& PersonaLive
& 0.681 & \textbf{0.782} & \textbf{0.051} & \textbf{1.056} & 439.3 \\
& PixReenact
& \textbf{0.728} & 0.802 & 0.074 & 1.318 & \textbf{432.2} \\

\bottomrule
\end{tabular}
\end{table*}

\paragraph{Extreme-condition breakdown.}
Table~\ref{tab:extreme_strata} breaks down the extreme-condition benchmark (Sec.~\ref{subsec:robust_streaming})
into extreme yaw, extreme pitch, accessory occlusion, and hand occlusion.
PixReenact consistently preserves the reference identity better than
PersonaLive across all four groups, while remaining competitive in motion and
temporal consistency.

\begin{table}[t]
\centering
\caption{\textbf{Extreme-condition benchmark by group.} Per-group
results of the streaming methods on the 68-clip extreme benchmark (Tab.~\ref{tab:extreme_benchmark}); mean\std{std} over clips.
FVD is omitted per group (set-level statistic, unstable at these set
sizes).}
\label{tab:extreme_strata}
\resizebox{0.7\textwidth}{!}{%
\begin{tabular}{lcccc}
\toprule
Method & ID$\uparrow$ & AED$\downarrow$ & APD$\downarrow$ & tLP$\downarrow$ \\
\midrule
\multicolumn{5}{c}{\textit{Extreme yaw (25 clips)}} \\
\midrule
PersonaLive & 0.334\std{0.20} & 1.416\std{0.46} & \textbf{0.178}\std{0.11} & 2.482\std{2.20} \\
PixReenact & \textbf{0.708}\std{0.09} & \textbf{1.302}\std{0.31} & 0.251\std{0.12} & \textbf{2.289}\std{1.89} \\
\midrule
\multicolumn{5}{c}{\textit{Extreme pitch (17 clips)}} \\
\midrule
PersonaLive & 0.566\std{0.18} & 1.342\std{0.64} & \textbf{0.115}\std{0.07} & \textbf{1.661}\std{1.00} \\
PixReenact & \textbf{0.733}\std{0.08} & \textbf{1.057}\std{0.15} & 0.126\std{0.06} & 1.954\std{0.90} \\
\midrule
\multicolumn{5}{c}{\textit{Accessory occlusion (16 clips)}} \\
\midrule
PersonaLive & 0.570\std{0.27} & \textbf{0.999}\std{0.22} & 0.096\std{0.07} & 1.416\std{0.76} \\
PixReenact & \textbf{0.728}\std{0.18} & 1.011\std{0.25} & \textbf{0.096}\std{0.05} & \textbf{1.396}\std{0.68} \\
\midrule
\multicolumn{5}{c}{\textit{Hand occlusion (10 clips)}} \\
\midrule
PersonaLive & 0.503\std{0.17} & 1.383\std{1.15} & \textbf{0.125}\std{0.11} & 1.395\std{0.66} \\
PixReenact & \textbf{0.761}\std{0.04} & \textbf{1.050}\std{0.32} & 0.152\std{0.11} & \textbf{1.321}\std{0.47} \\
\midrule
\multicolumn{5}{c}{\textit{All (68 clips)}} \\
\midrule
PersonaLive & 0.472\std{0.23} & 1.295\std{0.64} & \textbf{0.135}\std{0.10} & 1.866\std{1.57} \\
PixReenact & \textbf{0.727}\std{0.11} & \textbf{1.135}\std{0.30} & 0.169\std{0.12} & \textbf{1.853}\std{1.35} \\
\bottomrule
\end{tabular}}
\end{table}

\begin{figure*}[t]
    \centering
    \includegraphics[width=\textwidth]{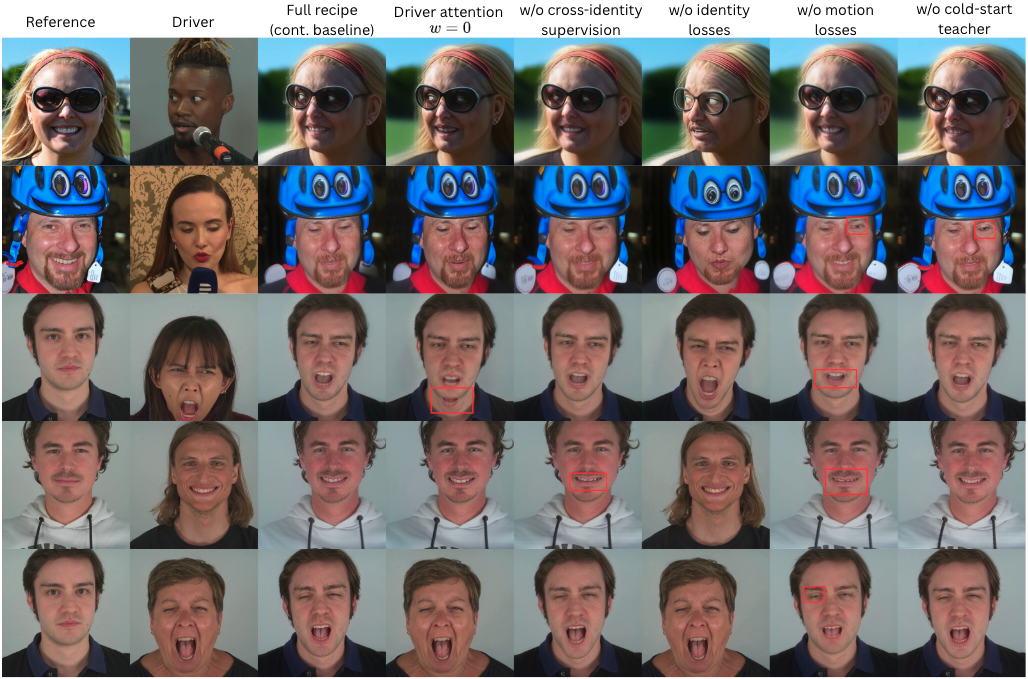} \caption{\textbf{Qualitative component ablations.} Representative cross-identity examples for the full recipe and its main ablations. Red rectangles highlight representative failure modes.}
    \label{fig:ablation_qual}
\end{figure*}

\paragraph{Qualitative ablations.}
Figure~\ref{fig:ablation_qual} complements the quantitative component
ablations in Table~\ref{tab:ablation_components}. Weaker identity constraints
and narrower driving attention can increase driver-appearance leakage, while
removing motion supervision reduces expression and pose transfer. Removing
cross-identity supervision or the cold-start teacher can also introduce local
artifacts and weaker motion transfer.

\paragraph{Additional ablations.}
Table~\ref{tab:ablation_additional} reports three additional sets of controlled
experiments. Their protocol differs from the main-paper ablations, where all
variants start from the same full-model checkpoint and are continued for 5K
steps with one component removed. Here, each group addresses a separate
training-design question and therefore uses its own matched setup. Results are
directly comparable only within the same group. For the \textbf{DMD ablation},
the variants with and without DMD are initialized from the same dense teacher
and trained for 20K steps. For the \textbf{pseudo-target filtering ablation},
the filtered and unfiltered variants start from the same pre-filter checkpoint
and are trained for an additional 5K steps, differing only in whether the
cross-identity pseudo targets are filtered. Finally, the
\textbf{self-reenactment ablations} follow the same continuation protocol as
the main-paper ablations: starting from the full model, we continue training for
5K steps after removing either the reconstruction losses or the real
self-driving data. The results show that DMD improves temporal consistency and
FVD, pseudo-target filtering provides modest gains in identity and motion
accuracy, and reconstruction losses and real self-driving samples improve
self-reenactment fidelity.

\begin{table*}[t]
\centering
\caption{\textbf{Additional ablations.} DMD variants are trained for 20K steps from the same dense teacher, and filtering variants for 5K steps from the same pre-filter checkpoint. Self-reenactment variants continue the full model for 5K steps. Values are comparable only within each block.}
\label{tab:ablation_additional}

\small
\setlength{\tabcolsep}{5pt}

\begin{tabular}{llcccccc}
\toprule
& Variant
& ID$\uparrow$ & IDD$\downarrow$ & AED$\downarrow$ & APD$\downarrow$
& tLP$\downarrow$ & FVD$\downarrow$ \\
\midrule
\multirow{2}{*}{\textit{DMD}}
& with DMD
& 0.618 & 0.029 & 0.944 & 0.155 & 0.967 & 508 \\
& w/o DMD
& 0.643 & 0.026 & 0.966 & 0.157 & 1.178 & 553 \\
\midrule
\multirow{2}{*}{\textit{Filtering}}
& filtered
& 0.841 & 0.018 & 0.977 & 0.161 & 0.995 & 630 \\
& unfiltered
& 0.819 & 0.019 & 0.997 & 0.167 & 1.001 & 609 \\
\bottomrule
\end{tabular}

\vspace{2pt}
\textbf{(a) Cross-identity training ablations}

\vspace{7pt}

\begin{tabular}{lcccccc}
\toprule
& \multicolumn{3}{c}{TalkingHead-1KH}
& \multicolumn{3}{c}{VFHQ-Test} \\
\cmidrule(lr){2-4}
\cmidrule(lr){5-7}
Variant
& PSNR$\uparrow$ & LPIPS$\downarrow$ & FVD$\downarrow$
& PSNR$\uparrow$ & LPIPS$\downarrow$ & FVD$\downarrow$ \\
\midrule
Full recipe
& 37.33 & 0.019 & 8.3
& 33.94 & 0.032 & 5.6 \\
w/o reconstruction losses
& 35.40 & 0.043 & 45.6
& 33.32 & 0.036 & 6.2 \\
w/o self-driving data
& 36.63 & 0.035 & 29.9
& 34.04 & 0.035 & 5.8 \\
\bottomrule
\end{tabular}

\vspace{2pt}
\textbf{(b) Self-reenactment fidelity}

\end{table*}

\end{document}